\documentclass[11pt]{article}
\usepackage[final]{acl}
\usepackage{fontspec}
\newfontfamily\cjkfont[Path=fonts/,ItalicFont=NotoSansCJKsc-Regular.otf,BoldFont=NotoSansCJKsc-Regular.otf]{NotoSansCJKsc-Regular.otf}
\newcommand{\zh}[1]{{\cjkfont #1}}
\newcommand{\ja}[1]{{\cjkfont #1}}
\newcommand{\ko}[1]{{\cjkfont #1}}
\usepackage{microtype}
\usepackage{amsmath,amssymb,graphicx,booktabs,tabularx,array,multirow}
\usepackage{placeins}
\newcommand{\md}{\lvert\Delta\rvert}
\title{Same Judgment, Different Yes:\\Measuring Multilingual Judgment Profiles in Language Models}
\author{Haonan Huang \\ Princeton University \And Yuwei Lu \\ Independent Researcher \And Chengkun Cai \\ University College London}
\begin{document}
\maketitle
\begin{abstract}
Multilingual evaluations often score a model's opening yes or no, a token that also carries model and language answering conventions.
We ask when different ways of eliciting and reading an answer recover the same judgment, and when they measure an answer habit.
We collect 1.1 million replies from eight models in six languages on up to 105 moral vignettes, varying the question's form, the answer constraint, and the reading of the reply, token or sentence, in both wordings (the FAR-6 battery).
Story-level judgment profiles converge across the tested readouts and question forms, except for the yes/no token after a negative interrogative read under a preset mapping: on the same replies, token and sentence profiles agree at a median $r=0.67$ across cells, while sentence profiles agree with free-sentence profiles at 0.94, a known linguistic variability.
With each language's story consensus removed, a model's profile is nearest to its own in another language for 39 of 48 model--language cells under proposition labels and 15 under free sentences, against 5 expected by chance.
Particle, position, and named-term preferences follow their own patterns.
A model can keep its judgment and change its yes: scoring that token requires establishing which proposition it affirms.
\end{abstract}

\section{Introduction}\label{sec:intro}
\begin{table*}[t]
\centering\footnotesize
\setlength{\tabcolsep}{2.1pt}
\newcommand{\yes}{\checkmark}
\newcommand{\no}{$\times$}
\begin{tabular}{@{}lcccccccccc@{}}
\toprule
 & Question & No & Both & Negative & Answer & Option & Repeated & Sentence & Same reply & Identity \\
 & languages & reference & wordings & question & formats & order & generations & judge & read twice & across \\
Study & & answer & & & & varied & & & & languages \\
\midrule
\citet{braun2025} & 3 & \no & \no & \yes & 2 & \no & \no & \no & \no & \no \\
\citet{moore2024} & 4 & \yes & \no & \no & 2 & \yes & \no & \yes & \no & \no \\
\citet{labat2026} & 8 & \yes & \no & \no & 1 & \yes & \no & \no & \no & \no \\
\citet{hua2025} & 1 & \no & \no & \no & 1 & \no & \no & \yes & \yes$^{a}$ & \no \\
\citet{okada2026} & 1 & \yes$^{b}$ & \no & \no & 2 & \yes & \no & \no & \no & \no \\
\citet{huang2026} & 1$^{c}$ & \yes & \yes & \no & 3 & \yes & \yes & \yes$^{d}$ & \no & \no \\
\citet{sha2026} & 2 & \yes & \no & \no & 1 & \no & \yes & \no & \no & \yes$^{e}$ \\
\midrule
This paper & 6 & \yes & \yes & \yes & 4 & \yes$^{f}$ & \yes & \yes & \yes & \yes \\
\bottomrule
\end{tabular}
\caption{Comparison with the closest measurement studies. Question languages: languages in which the questions are posed. Both wordings: each question also asked with the opposite evaluative term. Answer formats: elicitation formats compared on the same content. Option order varied: randomized \citep{moore2024}, randomized in the forced choice \citep{okada2026}, all permutations \citep{labat2026}, or crossed \citep{huang2026}. Same reply read twice: one reply read by a fixed rule and by a judge. Identity across languages: a model recognized across languages from its answer profile. $^{a}$Two of six benchmarks; twelve templates vary the output instructions; the multiple-choice rule scores option log-likelihoods, not the reply. $^{b}$Models answer as assigned personas. $^{c}$English questions; German, French, and Chinese yes/no words appear only as answer labels. $^{d}$On the free-choice replies; the forced binary is read by its label. $^{e}$Classifiers trained in one language, tested in the other. $^{f}$Paired within each item for proposition labels and yes/no wordings; German and French particle order rotates across items.}
\label{tab:related}
\end{table*}
 Language models are now asked for judgments, moral, social, and political, in many languages, and the usual instrument is a question with a fixed set of answers whose chosen option becomes a number: count yes as endorsement and no as rejection, then compare the number across models and languages \citep{moore2024,labat2026}, a number that is cheap, comparable, and looks language-neutral.

Nevertheless, the role of the language cannot be assumed to be null.
A classic case is the negative question: an answer can address the polarity of the question or the truth of the state of affairs asked about, and which of the two yes and no pick depends on the language and, within a language, on the construction.
To ``Do you not agree?'', an English speaker who does not agree may answer ``Yes'' or ``No'' and mean the same thing \citep{goodhue2018,holmberg2016}; a Japanese \ja{はい} (\emph{hai}, ``yes'') to a neutral negative question confirms the negative, that the speaker does not agree \citep{kuno1973}; German \emph{doch} and French \emph{si} exist to reject a negative antecedent \citep{claus2017,kameyama2013}.
The particle is an answering convention, not the judgment; a model trained on these languages inherits conventions of its own, such as preferring no, the second-listed option, or the term the question names.
A yes/no score therefore mixes the judgment with the way of answering, and the mixture can differ by language.
A model can keep its judgment and change its yes.

This is well described for human speakers but elusive in the evaluation of language models, where the form of the question, the constraint on the answer, and the rule that reads the reply are fixed by convenience rather than varied by design (\S\ref{sec:related}).
We ask when different ways of eliciting and reading an answer recover the same judgment, and when they measure an answer habit instead.
We distinguish a \emph{judgment profile}, the vector of story-level judgments of one model in one language under one readout, from an \emph{answer habit}, a systematic preference in the way of answering.
The battery (\S\ref{sec:battery}) crosses the form of the question, including both wordings and the negative interrogative, the constraint on the answer, and the reading of the reply on the same stories in six languages for eight models, and reads the same replies twice.

We report four findings.
(1) Story-level judgment profiles converge across readouts and question forms, except for the yes/no token after a negative interrogative, which disagrees with the sentence it opens on the same replies (\S\ref{sec:convergence}, Figure~\ref{fig:convergence}).
(2) The same model in another language is closer than another model in the same language, and its consensus-removed profile identifies it across languages for 39 of 48 cells (\S\ref{sec:identity}, Figure~\ref{fig:identity}).
(3) Preferences for no, for the second-listed option, and for the named term have patterns of their own by model and language (\S\ref{sec:habits}, Figure~\ref{fig:habits}).
(4) After a negative interrogative, the token maps to its own sentence by construction, language, and model (\S\ref{sec:negation}, Figure~\ref{fig:negation}).

The study is motivated by how humans answer, but it measures models: a judgment measure must allow the same judgment to come with a different yes, and scoring the token requires establishing which proposition the model uses it to affirm (\S\ref{sec:discussion}).

\section{Related work}\label{sec:related}
Table~\ref{tab:related} places the battery among the closest measurement studies: option order, labels, and prompt format move multiple-choice and benchmark answers \citep{zheng2024,pezeshkpour2024,sclar2024}, and survey answers carry label, ordering, and acquiescence biases of their own, the last probed with negated-agreement questions whose ambiguity is noted but not resolved \citep{dominguez2024,tjuatja2024,braun2025}.
Consistency studies vary paraphrase, topic, format, and translation \citep{moore2024}, option order, symbols, and language on human-translated survey items \citep{labat2026}, or the prompting language of moral-reasoning tests \citep{khandelwal2024,agarwal2024}; none crosses the interrogative construction with the evaluative keying, and each reads its outputs once.
On what a score reads, much of the prompt sensitivity measured on English benchmarks with reference answers comes from rule-based scoring rather than from the generated answers \citep{hua2025}; a crossed English design separates the stance on moral dilemmas from order and label effects across three instruments \citep{huang2026}; and a desirability-matched format mitigates instruction-induced socially desirable responding on English personality items \citep{okada2026}.
Model identity has been recovered by classifiers from psychometric profiles across two languages \citep{sha2026} and from generated text \citep{sun2025}.
The battery differs on three counts: question form and wording polarity are factors alongside the answer constraint in six languages; the same reply is read by a token rule and by a sentence judge; and identity is tested on consensus-removed profiles across six languages without a trained classifier.

\section{The battery and the measurement contract}\label{sec:battery}
\begin{figure*}[t!]
\centering
\includegraphics[width=.82\textwidth]{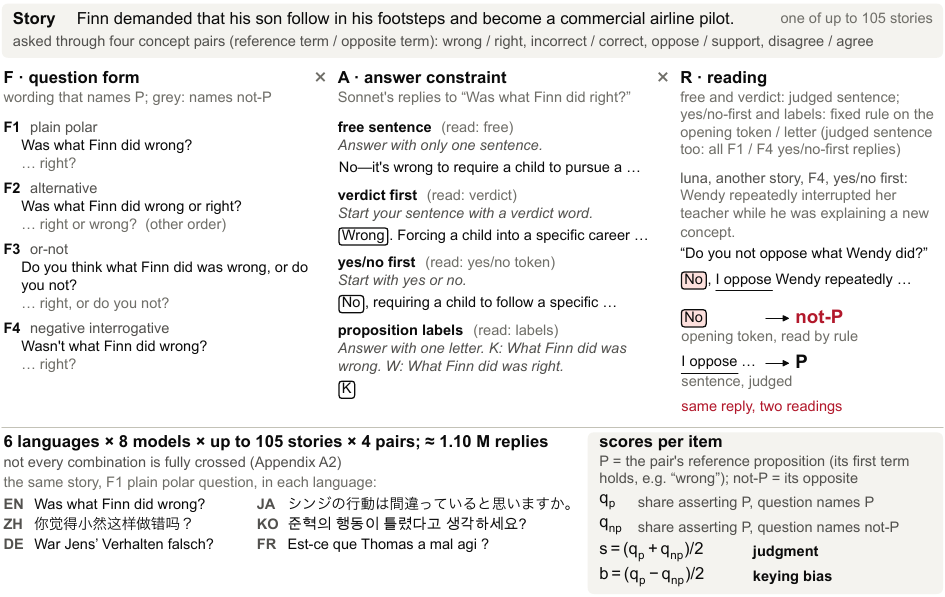}
\caption{The battery (FAR-6). One story is asked through concept pairs, question forms, and answer constraints, each question in both wordings, so that a judgment can be separated from a preference for a wording or a position. The reply is read twice, by a fixed decoder on its opening and by a sentence judge of the whole sentence; the example shows the two readings disagreeing on one reply to a negative interrogative. The form labels F1--F4 and the constraint names are used in the figures throughout.}
\label{fig:battery}
\end{figure*}
The battery (FAR-6; Figure~\ref{fig:battery}) varies three things about a yes/no question put to a model.
The first is the \emph{form} of the question, covering the general ways of posing one.
A plain polar question (``Was what Finn did wrong?'') is the default; the negative interrogative (``Wasn't what Finn did wrong?'') negates it, the construction on which answer systems differ; the alternative question lists both terms (``wrong or right?''); and the or-not question (``\ldots, or not?'') is natural in some of our languages and marked in others (Appendix~\ref{app:readouts}).
The second is the \emph{answer constraint}, the standard ways of requesting a reply, from the least to the most constrained: a free sentence, a sentence that starts with a verdict word, a sentence that starts with yes or no (with language-specific particles), and one letter that stands for an explicit proposition, with both letter orders presented (Appendix~\ref{app:protocol}).
The third is the \emph{reading}: the opening token or letter, read by a fixed rule, or the whole sentence, read by a judge; each form and each constraint is asked about the same stories in six languages and in both wordings of every question.

The stories combine 85 Moral Foundations Vignettes with published wrongness ratings \citep{clifford2015} and 20 authored stories, spanning acceptable, contested, and violation cases.
English, Chinese, German, and Japanese cover up to 105 stories, Korean and French 45.
Language-specific concept pairs such as wrong/right and oppose/support express opposing evaluations of the act (Appendix~\ref{app:pairs}); \emph{keying} denotes which term the question names.
Concept pair, keying, option order, and repetition (a subset of prompts receives three independent generations) stay separate throughout (Table~\ref{tab:battery}).

A \emph{readout} combines a question form, answer constraint, and reading; a \emph{cell} combines a model and a language.
The fixed token rule reads yes as affirming the term named in the question and no as its opposite, counting German \emph{doch} and French \emph{si} with yes.
After a negative interrogative this is the polarity-based mapping of the linguistic literature, one human answer system among others, which we retain as the hypothesis to be tested against the sentence it opens (\S\ref{sec:negation}).
The verdict word is read by a fixed rule extended in four stated ways (Appendix~\ref{app:parsing}; the figures label the rules R and R+).
The sentence judge (Claude Sonnet 5, one uniform prompt) instead codes whether the reply asserts the reference proposition, asserts its opposite, takes both sides or conditions, or takes no position (P and not-P in the figures); only the first two readings are decisive and enter the shares.

We orient both question wordings toward the same reference proposition before averaging, so that an answer to ``right?'' can be compared with an answer to ``wrong?''
Let $q_p$ and $q_{np}$ be the shares of decisive replies asserting the reference proposition when the question names that term and when it names its counterpart.
With $q_k=\mathrm{mean}_{\text{orders}}\Pr(\text{reference}\mid k,\text{decisive})$ for the term $k$ named in the question, judgment $s=(q_p+q_{np})/2$ averages the shares, whereas keying bias $b=(q_p-q_{np})/2$ measures their difference; repeated generations are averaged within the exact item before available orders receive equal weight.
Both wordings (or both orders of an alternative question) must have decisive support, and the story score averages $s$ over concept pairs.

We compare profiles on their common stories through three quantities: \emph{story-level $r$}, their Pearson correlation; \emph{residual-profile $r$}, the same correlation after removing each language's story consensus over models and each model's mean, which asks whether a model's departures from its peers stay distinctive; and \emph{mean $\md$}, the mean absolute difference between corresponding story scores, which retains changes in individual judgments that a high correlation can conceal.
Class means use 95\% story-bootstrap intervals with models and languages fixed, and class differences the same resampled stories.

The battery holds 1,104,574 replies, 839,230 of them in the analyzed answer constraints; the rest are additional conditions and a control stratum excluded from all analyses, and not every combination of form and constraint is run.
We evaluate Claude Haiku 4.5, Claude Sonnet 5, Claude Opus 5, GPT-6 luna, GPT-6.1 sol, Gemini 3.1 Flash-Lite, Qwen3.8-27B, and Mistral Medium 3.5 (prompts, settings, and coverage in Appendices~\ref{app:protocol} and \ref{app:coverage}).
Every free and verdict-first reply has a judged sentence, as does every yes/no-first reply to the plain question and to the negative interrogative; about 4\% of the other token and label replies are judged (Appendix~\ref{app:coverage}).
Two other judge models agree with the sentence reference on 95.5\% and 97.2\% of audited replies, native readers on 95.5\% of 400 replies in four languages with no opposite decisive label, and repeated runs of the same judge on 96.3\% and 97.7\% of two sets of 300 identical prompts (Table~\ref{tab:audits}).

Two readouts of one cell can differ for three reasons: sampling variability across repeated generations, which more draws reduce (Appendix~\ref{app:stability}); systematic effects of wording, keying, order, and constraint, which the balanced design averages over and \S\ref{sec:habits} measures one by one; and the mapping from an opening token to a proposition, which no amount of sampling can settle, because the same yes accompanies opposite propositions in different constructions.
Reading the same replies twice isolates the third: both readings of the same replies with identical weights hold the generated text fixed, so a difference lies in the reading; comparing those sentences with free sentences on the same stories then checks whether the constraint changed the profile.

\begin{figure*}[t!]
\centering
\includegraphics[width=.80\textwidth]{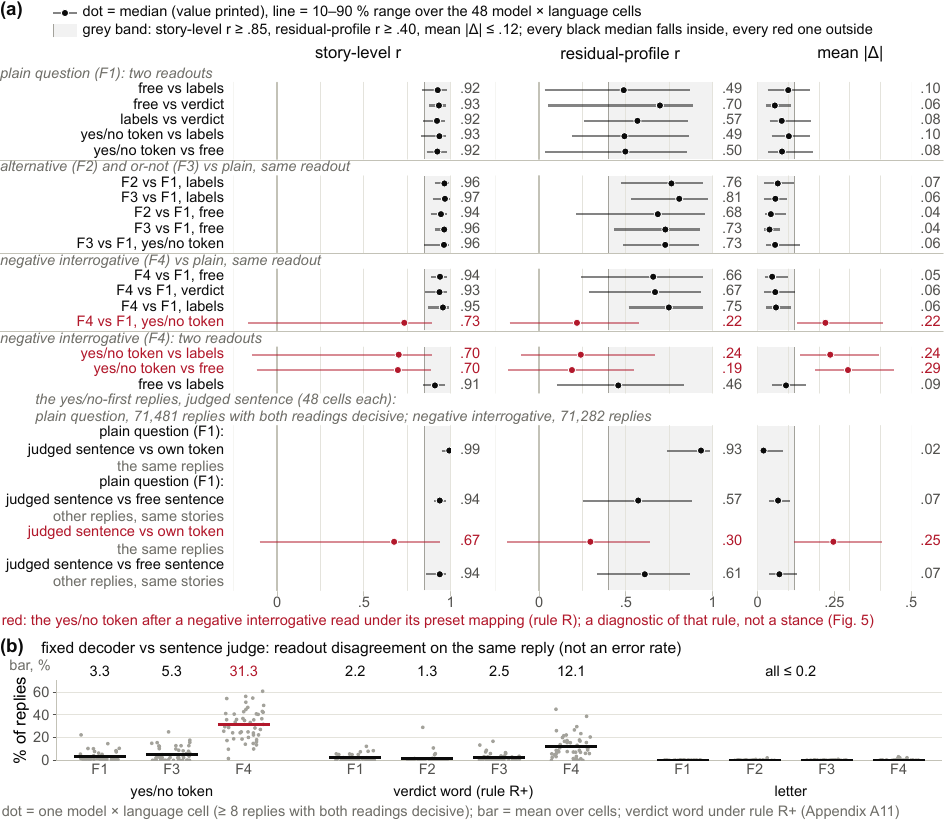}
\caption{Which readouts converge. (a) Medians and 10--90\% ranges across cells for each readout pair, grouped by question form, with a pale band over the convergent range; red marks the negative-interrogative token under its preset mapping. The last four rows read the same yes/no-first replies twice, sentence against own token, for the plain question (71,481 replies) and the negative interrogative (71,282), and compare those sentences with free sentences of other replies on the same stories. (b) Disagreement between the fixed decoder and the sentence judge on the same reply, one dot per cell and the bar at the cell mean; verdict words use the extended rule (Appendix~\ref{app:parsing}). Reply counts per series are in Appendix~\ref{app:parsing}.}
\label{fig:convergence}
\end{figure*}

\section{Which readouts converge}\label{sec:convergence}
Figure~\ref{fig:convergence}a gives each comparison's median over the 48 cells and its 10--90\% range on the stories common to the compared readouts (Appendix~\ref{app:readouts}).

The plotted readout pairs converge except for those red rows, with median story-level $r$ at least 0.90.
Under plain questions, free sentences agree with proposition labels at 0.92 and with verdict-first sentences at 0.93; the comparisons across question forms preserve the profile likewise (Table~\ref{tab:readouts}).
Median residual correlations among the convergent pairs range from 0.46 to 0.93: agreement retains how individual models depart from the consensus.

Under the fixed rule, the negative-interrogative token changes that picture: against the plain-question token its median story-level correlation falls to 0.73, with mean absolute story difference 0.22 and some cells below zero.
On the same 71,282 replies with both readings decisive (\S\ref{sec:battery}), the token and sentence profiles agree at 0.67, with absolute difference 0.25, whereas the profiles of those judged sentences agree with free-sentence profiles from other replies on common stories at 0.94 (absolute difference 0.07); the same double reading of the plain-question replies agrees at 0.99 with absolute difference 0.02.
Requiring a yes/no opening still produces sentences whose story profiles reproduce the free-sentence profile; the change lies in the opening's mapping to its sentence.

Reply-level disagreement supports this reading without aggregation (Figure~\ref{fig:convergence}b): among replies with both readings decisive, mean disagreement across cells rises from 3.3\% for plain-question yes/no-first replies to 31.3\% after a negative interrogative.
A targeted audit by three judge configurations at high reasoning effort upholds the sentence reading in 93--96\% of the conflicting negative-interrogative replies but in only about half of the conflicting plain-question replies (Appendix~\ref{app:protocol}), so the plain-question rate partly reflects the judge.
Verdict openings also depart more under negation; proposition letters do not (Appendix~\ref{app:parsing}).

\begin{figure*}[t!]
\centering
\includegraphics[width=.83\textwidth]{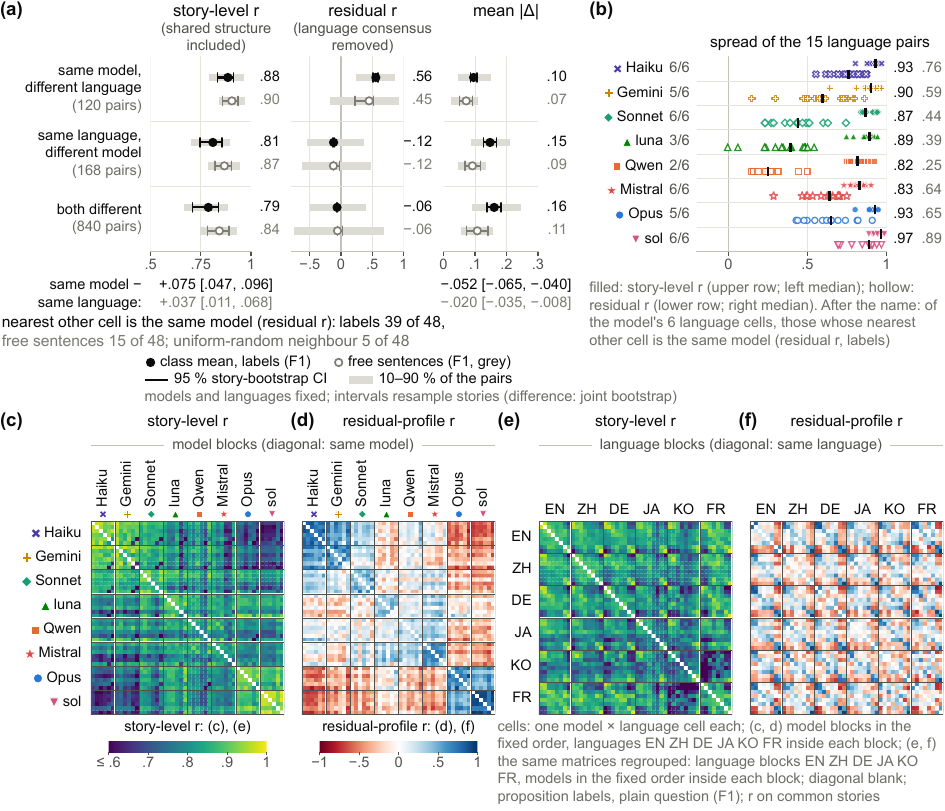}
\caption{Model identity across languages. (a) Class means with 95\% story-bootstrap intervals and pair ranges, labels filled and free sentences hollow, with the nearest-neighbor counts. (b) Per-model language-pair values, filled for story-level $r$ and hollow for residual $r$, with the number of the model's six language cells whose nearest other cell is the same model (residual $r$, labels). (c) Between-cell story-level correlations on common stories and (d) residual-profile correlations, cells grouped by model; (e) and (f) the same matrices grouped by language: the same-model blocks on the diagonal of (d) stand out, the same-language blocks on the diagonal of (f) do not; the counts in (a) and (b), not the heat maps, identify a model.}
\label{fig:identity}
\end{figure*}

\begin{figure*}[t!]
\centering
\includegraphics[width=.76\textwidth]{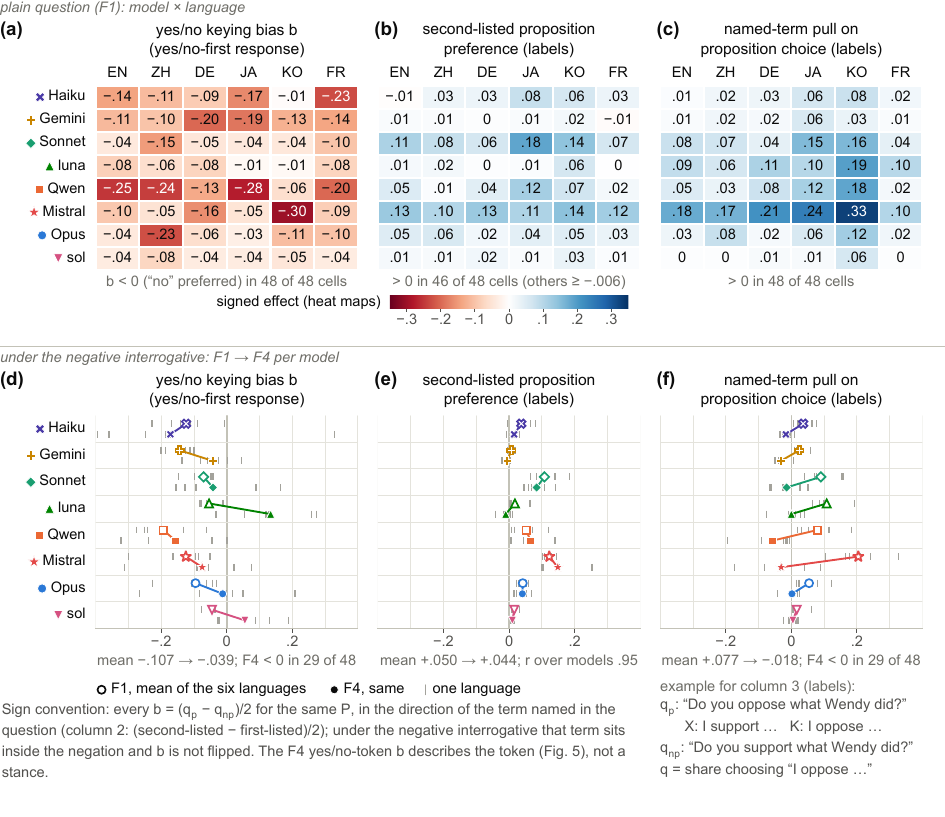}
\caption{Three answer habits under plain questions (top) and negative interrogatives (bottom). Model means over languages use hollow markers for plain questions and filled markers for negative interrogatives; ticks mark languages. The second-listed preference remains similar, named-term pull approaches zero, and token keying bias scatters and changes sign.}
\label{fig:habits}
\end{figure*}

\section{Model identity across languages}\label{sec:identity}
With the readouts agreeing except where noted, profiles can be compared across models and languages: is the same model in another language closer than another model in the same language?
Figure~\ref{fig:identity} compares proposition-label profiles under plain questions over pairs of model--language cells, grouped by what changes: language, model, or both.

The most direct test treats the 48 cells (eight models in six languages) as unlabeled profiles and asks which other cell each one resembles most (Table~\ref{tab:pairs}).
Under residual $r$, a cell's nearest neighbor is another language of the same model for 39 of 48 proposition-label cells (per model in Figure~\ref{fig:identity}b), against 5 under a uniform-random-neighbor benchmark that gives each of the 47 other cells equal probability and ignores the language structure of the cells; under story-level $r$ the count is 25, and under mean $\md$ it is 30.

The same-model class has higher mean story-level correlation than the same-language class, 0.88 against 0.81, and a mean absolute story difference of 0.10 against 0.15; both contrasts have joint bootstrap intervals that exclude zero (\ensuremath{+0.075} [0.047, 0.096] and \ensuremath{-0.052} [\ensuremath{-0.065}, \ensuremath{-0.040}]).
Story-level $r$ is dominated by structure every cell shares (violations high, acceptable stories low), which keeps the gap modest.
Within one story group the gap widens where models disagree, 0.79 against 0.69 on acceptable stories and 0.70 against 0.62 on contested stories, and vanishes on violations (0.43 against 0.46), where every cell condemns and little remains to tell models apart (Appendix~\ref{app:means}).

The same model across languages retains a mean residual correlation of 0.56; within-language centering couples the different-model residuals, so their negative mean is a constrained reference, and the informative result is the positive same-model association.

The strength of that signature differs across models and language pairs (Figure~\ref{fig:identity}b).
Residual correlations span 0.69--0.97 for sol against 0.14--0.50 for Qwen and about zero to 0.54 for luna.
Under free sentences the residual profiles identify the model for 15 of 48 cells, against 39 under proposition labels.
One candidate explanation is granularity: free-sentence story scores sit at an endpoint for 81\% of model--language--story observations against 49\% for labels (Appendix~\ref{app:stability}), leaving less graded structure to carry a model's departures; the comparison does not establish this.
A model thus retains a distinctive judgment profile across languages, with story-level changes of 0.10 (labels) and 0.07 (free sentences) per story.

\begin{figure*}[t!]
\centering
\includegraphics[width=.80\textwidth]{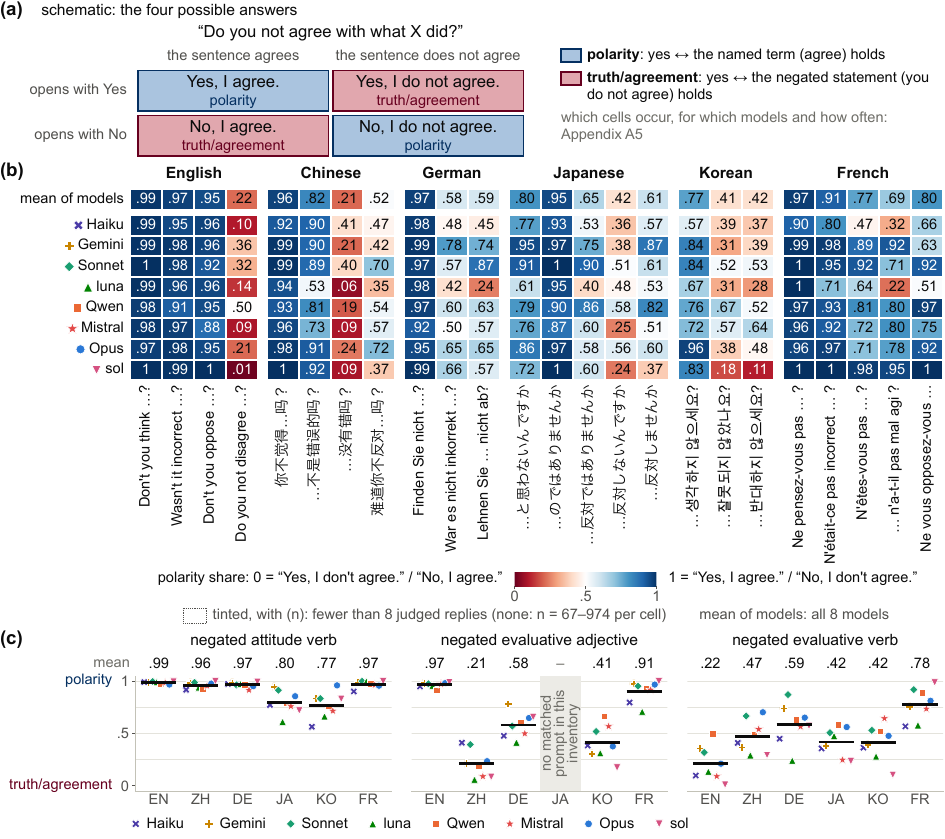}
\caption{The negative-interrogative token, given the sentence. (a) Particle--sentence combinations define the mappings. (b) Polarity shares by wording and model: blue is polarity, red truth/agreement, white an even split. (c) Corresponding constructions across languages (negated attitude verb, negated evaluative adjective, negated evaluative verb; wordings in Table~\ref{tab:constructions}); bars give model means and markers individual models. Appendix~\ref{app:negation} gives the direction split.}
\label{fig:negation}
\end{figure*}

\section{Answer habits and their change under negation}\label{sec:habits}
We now return to the constrained readouts, the yes/no token under the plain question and the proposition letter, whose story profiles converge with the sentence profile while measurable answering preferences remain, attached to the question's wording or option order rather than to the story.
We measure three such habits on the same reference proposition.
Yes/no keying bias is $b$, half the difference between the shares under the two wordings (negative for a preference for no; the sign is kept when the named term sits inside negation).
Second-listed preference, measured on proposition-label choices, is half the difference between choosing the second- and the first-listed proposition.
Named-term pull, a term of ours for the same choices, is the preference for the proposition that contains the term the question names, a lexical echo of the question (Figure~\ref{fig:habits}).

Yes/no keying bias favors no in every plain-question cell, averaging \ensuremath{-0.11}, but its magnitude belongs to a model--language combination: the interaction accounts for 0.56 of the cell-mean matrix's variance (cells in Appendix~\ref{app:habits}).

Second-listed preference averages \ensuremath{+0.05} and follows the model more closely; named-term pull is positive in every plain-question cell, averaging \ensuremath{+0.08} (Table~\ref{tab:habits}).
Position balance and opposing wordings therefore address distinct preferences: reversing the options alone leaves the named-term association free to move the answer (Appendix~\ref{app:habits}).

Negation separates the habits' trajectories (hollow against filled markers): second-listed preference remains similar, named-term pull approaches zero, and the fixed reading of the yes/no opening changes the sign of $b$ in 19 of the 48 cells.
The changing sign points to the token's mapping to its sentence.

\section{What the negative-interrogative token encodes}\label{sec:negation}
Figure~\ref{fig:negation}a illustrates the alternatives for ``Do you not agree with what X did?'': under the \emph{polarity mapping}, yes accompanies ``I agree''; under the \emph{truth/agreement mapping}, yes accompanies ``I do not agree.''
The \emph{polarity share} is the fraction of usable particle replies following the former mapping given their judged sentence (selection in Appendix~\ref{app:coverage}).

Judged free sentences under negative interrogatives have median story-level correlation 0.94 with their plain-question counterparts, and against the published English human reference the token's mean correlation falls from 0.60 to 0.25 while the sentence and label associations remain similar (Appendix~\ref{app:human}).

For Japanese \ja{反対しないんですか} (\emph{hantai shinai n desu ka}, ``Do you not oppose?''), Sonnet, Opus, and Qwen favor polarity, while Haiku, luna, and Mistral favor truth/agreement.
In English, ``Do you not agree / oppose?'' has mean polarity share 0.22 against 0.99 for ``Don't you think?'' (Appendix~\ref{app:negation}).

Families of corresponding constructions, defined from the prompts before inspecting their values (Figure~\ref{fig:negation}c), show the negated evaluative adjective ranging from mean polarity shares 0.97 in English and 0.91 in French to 0.21 in Chinese (wordings and support in Table~\ref{tab:constructions}).

Human linguistics distinguishes a few answer systems for negative questions: polarity-based ones, in which yes affirms the term the question names; truth- or agreement-based ones, in which yes confirms the negative proposition; and mixed ones with a dedicated reversal particle such as German \emph{doch} or French \emph{si} \citep{holmberg2016,claus2017,kameyama2013}.
English is usually placed in the first group and Japanese in the second \citep{kuno1973}, and in Mandarin \zh{是}/\zh{不} (\emph{shì}/\emph{bù}) track the answerer's opinion against the polarity of the question's main point \citep{lu2002}; our two mappings operationalize the first two systems.
Speakers are less clear-cut than the classification, with both answers attested in one language depending on construction and context \citep{goodhue2018,park2023}; the models show the same complexity.
Appendix~\ref{app:background} sets the model mappings beside the literature and dialogue corpora; a matched human study is future work.
We therefore identify the mapping from the model's own sentence: a changed yes can reflect a changed way of relating to a negative question while the sentence preserves the judgment.

\section{Discussion and recommendations}\label{sec:discussion}
Read the sentence, or balance proposition labels across opposing wordings and orders; both reproduce the story profile.
Report keying bias beside a yes/no score; the preference for no varies by model and language.
Do not score the opening token after a negative interrogative as a judgment.
A judgment measure must allow the same judgment to come with a different yes.

\FloatBarrier
\section*{Limitations}
\paragraph{Profile-level convergence.}
Same-model comparisons retain per-story differences, and residual structure reproduces with substantial variation across models and language pairs; the model signature is sharper under proposition labels than under free sentences.
Medians over cells for readout comparisons and class means over cell pairs for identity comparisons summarize different units.
\paragraph{Coverage and sampling.}
Korean and French contain fewer stories.
Undecided sentences that take both sides, add conditions, or take no position are excluded from decisive shares, and the other token and label conditions are judged on a sample.
The English, Chinese, German, and Japanese instruments were reviewed by native readers; the Korean and French instruments and story renderings were not (Appendix~\ref{app:protocol}).
\paragraph{Sentence reference.}
The sentence judge is one model, supported by two other judge models, a targeted audit at higher reasoning effort, test--retest checks, and native-reader labels on 100 replies per language in Chinese, English, German, and Japanese; Korean and French replies have no human labels, and the audited negative-interrogative yes/no-first strata hold 19--33 replies per language.
These audits measure agreement between judgment procedures, leaving shared labeling biases possible; on plain-question replies where token and sentence conflict, stronger judges side with the token about half the time.
\paragraph{Population and protocols.}
The models and languages are fixed, and intervals resample stories rather than model or language populations.
The measurements concern the forms, constraints, reading rules, and wordings represented in the battery.
The construction groups retain available language-specific wordings and therefore do not isolate syntax from every lexical or pragmatic difference.
\paragraph{Human context.}
Published MFV ratings provide an English reference set used across languages.
The dialogue corpora span different registers and collection settings from the model prompts and provide linguistic background, without establishing that models match or mismatch speakers; no human study on the same stories exists.
\paragraph{Interpretation.}
$s$ and $b$ are an operational decomposition, not pure judgment and pure bias; the judge is a reference rather than ground truth; and cross-language identifiability is a property of the measured profiles, not evidence of a language-free internal state.

\bibliography{refs}
\clearpage
\appendix
\setcounter{section}{0}
\renewcommand{\thesection}{A\arabic{section}}
\setcounter{figure}{0}
\renewcommand{\thefigure}{A\arabic{figure}}

\section{Prompt construction, model settings, and judging}
\label{app:protocol}
\setcounter{table}{1}
\begin{table}[t]
\centering
\small
\setlength{\tabcolsep}{3pt}
\begin{tabularx}{\columnwidth}{@{}p{.24\columnwidth}>{\raggedright\arraybackslash}X@{}}
\toprule
Factor & Levels or support \\
\midrule
Stories & Up to 105 per language; Korean and French: 45. \\
Concept pairs & 4 per language; Japanese: 3; both keyings balanced. \\
Form & Plain polar; alternative; or-not; negative interrogative. \\
Constraint & Free sentence; verdict first; yes/no first; proposition labels. \\
Reading & Fixed decoder; judged sentence. \\
Languages & English, Chinese, German, Japanese, Korean, French. \\
Models & 8 models; 48 model $\times$ language cells. \\
Repetition & 3 independent generations per repeated prompt. \\
Total replies & 1\,104\,574; four analyzed constraints: 839\,230. \\
Decisive coverage & Free sentence 85\%; verdict sentence 94\%; yes/no token 99\%; proposition labels 99\%. \\
\bottomrule
\end{tabularx}
\caption{The battery (FAR-6). Coverage pools decisive replies over generated replies in the analyzed constraints; the yes/no-token denominator excludes alternative questions, whose token is unreadable by design. Combinations have different supports.}
\label{tab:battery}
\end{table}
 
\paragraph{Stories and prompts.}
The 105 stories are 85 Moral Foundations Vignettes, including 16 social-norm controls, and 20 authored stories. The three story groups (acceptable 26, contested 46, violation 33) were fixed before any model was run, from the published wrongness ratings and the authored design; Korean and French use 45 stories, 15 per group. Each story and language has four machine renderings (two translation services and two language models), of which two enter the battery and rotate across the story's two concept-pair blocks; an automatic screen (name, length, script, back-translation, cross-engine outlier) flagged 184 renderings, and 71 were held out after a reading. Native readers reviewed the English, Chinese, and German instruments (an author) and the Japanese instrument (a collaborator); the Korean and French instruments and renderings had no native review. Question templates and instructions were written per language, never translated. The design records each item's story group, language, rendering, concept pair, question form, named term, answer constraint, option order, and repeat, with the exact prompt text and the letter or particle mapping. Figure~\ref{fig:a7} gives the language-specific evaluative terms, while Figure~\ref{fig:negation} and Table~\ref{tab:constructions} give the negative-interrogative wordings.

\paragraph{Judged coverage.}
Every free and verdict-first reply has a judged sentence, as does every yes/no-first reply to the plain question and to the negative interrogative (77,952 each). For the other yes/no-first and proposition-label conditions a seeded sample is judged; Table~\ref{tab:audits} gives the counts per form.

\paragraph{Estimation steps.}
Shares are computed per item over decisive replies, averaged over repeated generations within the item, then over option orders with equal weight, combined over the two keyings into $s$ and $b$, and averaged over concept pairs into story scores. Two cells are compared on their common stories when at least eight exist. Residual profiles subtract, within each language, the story mean over models and then each cell's mean, and are compared on stories with common-model support; a cell without variance on the common support is skipped. Nearest neighbors are taken among the 47 other cells; no ties occurred in the reported readouts. Class intervals come from 300 bootstrap resamples of stories with a fixed seed, every cell recomputed per draw including the residual centering; class differences use the same draws.

\paragraph{Construction choices.}
The inventory distinguishes constructions because response interpretation depends on negation placement and context as well as language \citep{romero2004,goodhue2018,espinal2019}. Japanese descriptions distinguish neutral negative questions from questions with a positive expectation \citep{kuno1973,kameyama2013}. Korean studies distinguish short and long negation and show contextual effects on the acceptability of particle--sentence combinations \citep{wee2019,kim2024}; the Korean templates here cover long negation. These distinctions motivate retaining the wording and response direction in the analysis of the negative interrogative.

Mandarin descriptions discuss the \zh{是否}/\zh{是不是} (\emph{shìfǒu}/\emph{shì-bu-shì}) question frames \citep{lu1985} and how \zh{是}/\zh{不} (\emph{shì}/\emph{bù}) track the answerer's opinion against the polarity of the question's main point \citep{lu2002}, while elicited rejecting responses can combine particles, sentence content, prosody, and gesture \citep{li2016}. French \emph{oui}/\emph{non}/\emph{si} interpretation is sensitive to the scope of negation \citep{pasquereau2020}. Human survey studies also distinguish the meaning of an attitude from how respondents map it to yes/no options under contrasting evaluative wording \citep{holleman2006,kamoen2017}. This motivates balancing which of the concept pair's terms a question names as a separate design factor.

\paragraph{Answer instructions.}
The free condition asks for one sentence. The verdict-first instruction is ``Start your sentence with a verdict word.'' The yes/no-first instruction is ``Start with yes or no.'', with the reverse particle order tested where available. Proposition-label prompts ask ``Answer with one letter.'' and attach each letter to a complete proposition, such as ``Z: I oppose what Elsa did. W: I support what Elsa did.'' The design records the actual letters and their proposition map for each item.

German particle choices include \emph{ja}, \emph{nein}, and \emph{doch}; French choices include \emph{oui}, \emph{non}, and \emph{si}. Their order varies across items as recorded in the design. Additional battery conditions ask for a listed evaluative term, an English \emph{yes} or \emph{no} opening in a Chinese, German, or Japanese prompt, or a Chinese \zh{对}/\zh{不对} (\emph{duì}/\emph{bù duì}, ``correct/not correct'') opening. These conditions contribute to the battery's total reply count; the profile analyses use the answer constraints specified in the main text.

\paragraph{Generation settings.}
We evaluate Claude Haiku 4.5 (Haiku), Claude Sonnet 5 (Sonnet), Claude Opus 5 (Opus), GPT-6 luna (luna), GPT-6.1 sol (sol), Gemini 3.1 Flash-Lite (Gemini), Qwen3.8-27B (Qwen), and Mistral Medium 3.5 (Mistral). The design assigns the same prompt inventory to every model, with independent generations of the prompts selected for repetition. Haiku, Sonnet, and Opus use the non-thinking setting; luna uses no reasoning and sol uses low reasoning effort. Gemini and Mistral use their default generation settings, and Qwen runs with thinking disabled. These settings form part of the measured model, and Appendix~\ref{app:stability} reports their observed sampling variability.

\paragraph{Sentence judging.}
The judge sees the story, instruction, question, pair-specific reference proposition, and verbatim reply. Its task is to code what the reply asserts about that proposition while ignoring the opening token and its own evaluation of the story. The judge's categories distinguish sentences that assert the reference proposition, assert its opposite, take both sides or conditions, or take no position. The last category includes refusals and comments about the question that do not express a judgment. Only sentences asserting the reference proposition or its opposite receive a decisive reading. The fixed decoder separately reads the opening expression from the reply text and the prompt's recorded mapping.

The primary sentence labels use Claude Sonnet 5 with low adaptive reasoning and a uniform codebook. The cross-model audit compares those labels with GPT-6 astra at high reasoning effort and obtains agreement of $95.5\%$ on 4,590 replies. A separate comparison with Claude Fable 5 at high reasoning effort gives $97.2\%$ on 4,530 replies (60 of the sampled replies were refused by that judge's safety classifier and carry no label). The audited replies cover 153 strata of model, constraint group, and language, with 720--840 replies per language; 1,225 are negative-interrogative replies, among them 152 yes/no-first replies and 47 on which the fixed rule and the judged sentence disagree. Table~\ref{tab:audits} gives agreement and opposite-decisive counts per stratum, the intervals, and the counts on which the primary judge gave a decisive reading; the two audit judges reverse a decisive stance on 10 and 9 replies in all, and their other disagreements involve at least one undecided category. Native readers (the authors for Chinese, English, and German; a collaborator for Japanese) labelled 100 replies per language drawn from the same sample, stratified over judged part and constraint group, with the same categories and without model names or machine labels. They agree with the primary labels on 95.5\% of the 400 replies ($\kappa=0.89$; 96, 93, 96, and 97 of 100 in Chinese, English, German, and Japanese), with the astra labels on 93.0\%, and with the Fable labels on 95.4\% of 389 replies. On the 363 replies where reader and primary judge are both decisive, no label opposes the other, and each of the 18 differences is a reply the judge codes as taking both sides or conditions, typically a reply that takes a side and adds a reservation, which the readers code on that side.

Two test--retest sets repeat the same sentence judge on 300 identical negative-interrogative yes/no-first prompts each, 50 per language. Agreement is 96.3\% and 97.7\% ($\kappa=0.92$ and 0.95), with most differences involving undecided categories and one and 3 flips between opposite decisive judgments. These checks complement the cross-model audits by measuring repeatability under unchanged judging instructions.

A core-strata audit then re-judges, with Claude Fable 5, Claude Opus 5.5, and Claude Sonnet 5 at high reasoning effort, a seeded sample of 3,263 replies on which the fixed rule and the primary label are both decisive, drawn per model--language cell and split by whether the two readings agree or conflict (Table~\ref{tab:audits}). Where the readings agree, the audit judges agree with the primary label on 96--99\% of the replies. On the negative-interrogative yes/no-first replies where they conflict, the audit judges uphold the sentence reading against the token rule for 93.5\% (Fable 5), 92.8\% (Opus 5.5), and 96.4\% (Sonnet 5 high) of the replies, all three together for 809 of 880 replies with three labels and none of them for 23; on the verdict-first conflicts under the extended rule they uphold it for 91.0\%, 90.7\%, and 90.9\%. On the rarer plain-question conflicts they uphold it for only 48.7\%, 49.0\%, and 48.1\%, while agreeing with one another on 96.7--97.9\% of all core replies: about half of the plain-question disagreement in Figure~\ref{fig:convergence}b reflects the primary judge rather than the token, whereas the negative-interrogative disagreement does not. Fable 5 refused 3 of its 222 calls (45 replies) and Sonnet 5 at high effort exhausted its token budget on 14 calls (206 replies); those replies carry no label from that judge.

\setcounter{figure}{1}

\section{Coverage and selection}
\label{app:coverage}

\begin{figure*}[p]
\centering
\includegraphics[width=\textwidth,height=.84\textheight,keepaspectratio]{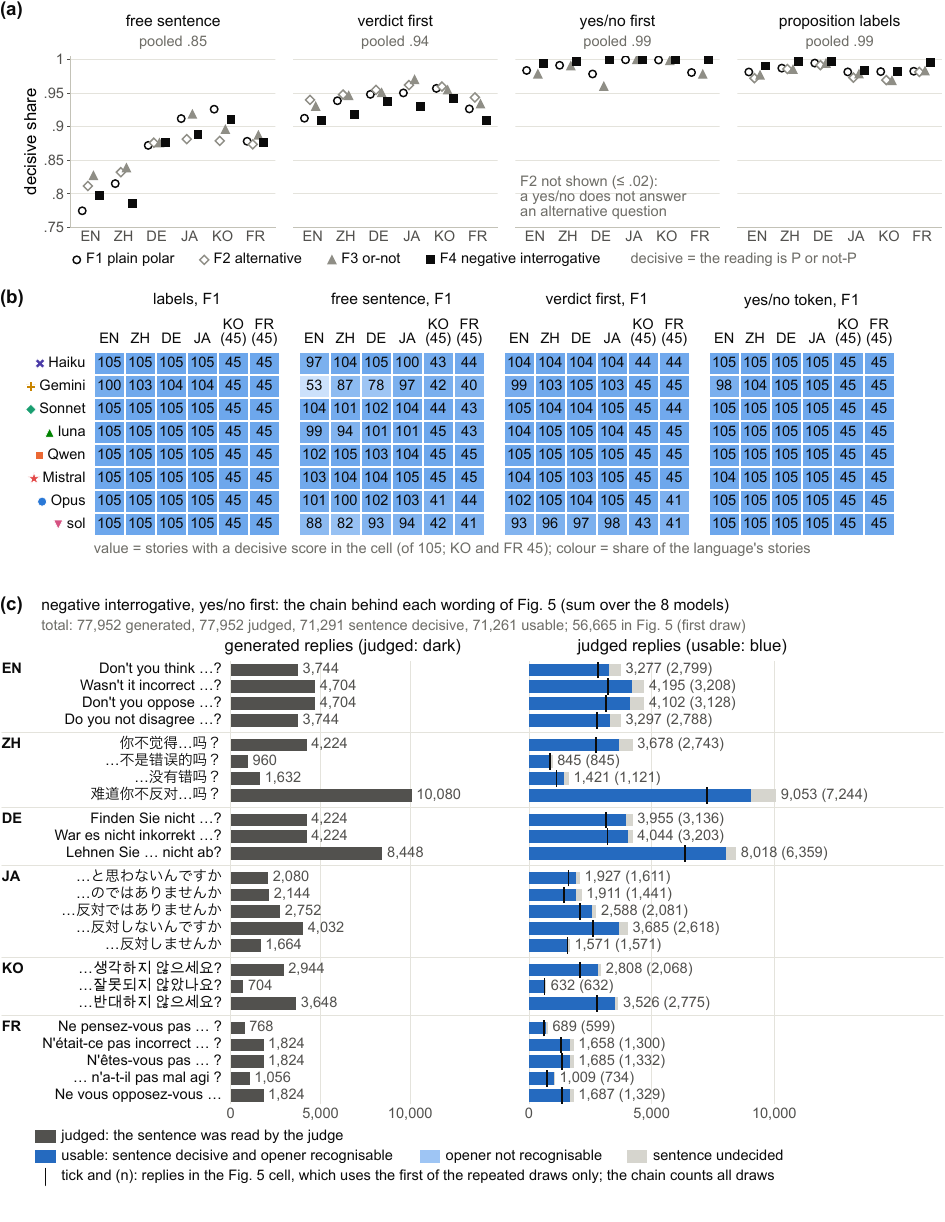}
\caption{Coverage and selection. Decisive coverage per answer constraint $\times$ form $\times$ language; stories per cell; and the chain behind every negative-interrogative yes/no cell of Figure~\ref{fig:negation}: generated $\rightarrow$ judged $\rightarrow$ sentence decisive $\rightarrow$ opener recognisable $\rightarrow$ usable. The chain counts all generations; the wording cells use the first generation of each prompt.}
\label{fig:a1}
\end{figure*}

Figure~\ref{fig:a1} separates decisive coverage, story support, and the selection of replies for the negative-interrogative mapping analysis. Decisive coverage uses the judged sentence for free and verdict-first answers, and the fixed decoder for yes/no-first and proposition-label answers. A reply contributes to a judgment profile only when the selected reading asserts the reference proposition or its opposite. Sentences taking both sides or conditions, sentences taking no position, and unreadable openings provide no decisive value for their respective readings. The alternative-question yes/no-first condition is excluded from these coverage summaries because a bare particle does not select between the alternatives.

The selection chain counts every generated reply of the negative-interrogative yes/no-first condition, including repeated generations. It contains 77,952 generated replies, all with a judged sentence. A reply enters the mapping analysis when its sentence is decisive and its opening is a recognised yes/no or reversal particle. Restricting this usable pool to the first generation of each prompt gives the wording cells in Figure~\ref{fig:negation}; this selection contains 56,665 replies.

The same-reply decoder diagnostic and the particle-mapping analysis have different inclusion rules. The former accepts any opening that the fixed decoder maps to a decisive reading in the yes/no-first condition. The latter additionally requires an opening particle so that the polarity share is defined. Figure~\ref{fig:a1} makes this particle-specific selection visible, while Table~\ref{tab:audits} reports the reply counts for the decoder comparison.

\section{Readouts and forms in detail}
\label{app:readouts}

\begin{figure*}[p]
\centering
\includegraphics[width=\textwidth,height=.84\textheight,keepaspectratio]{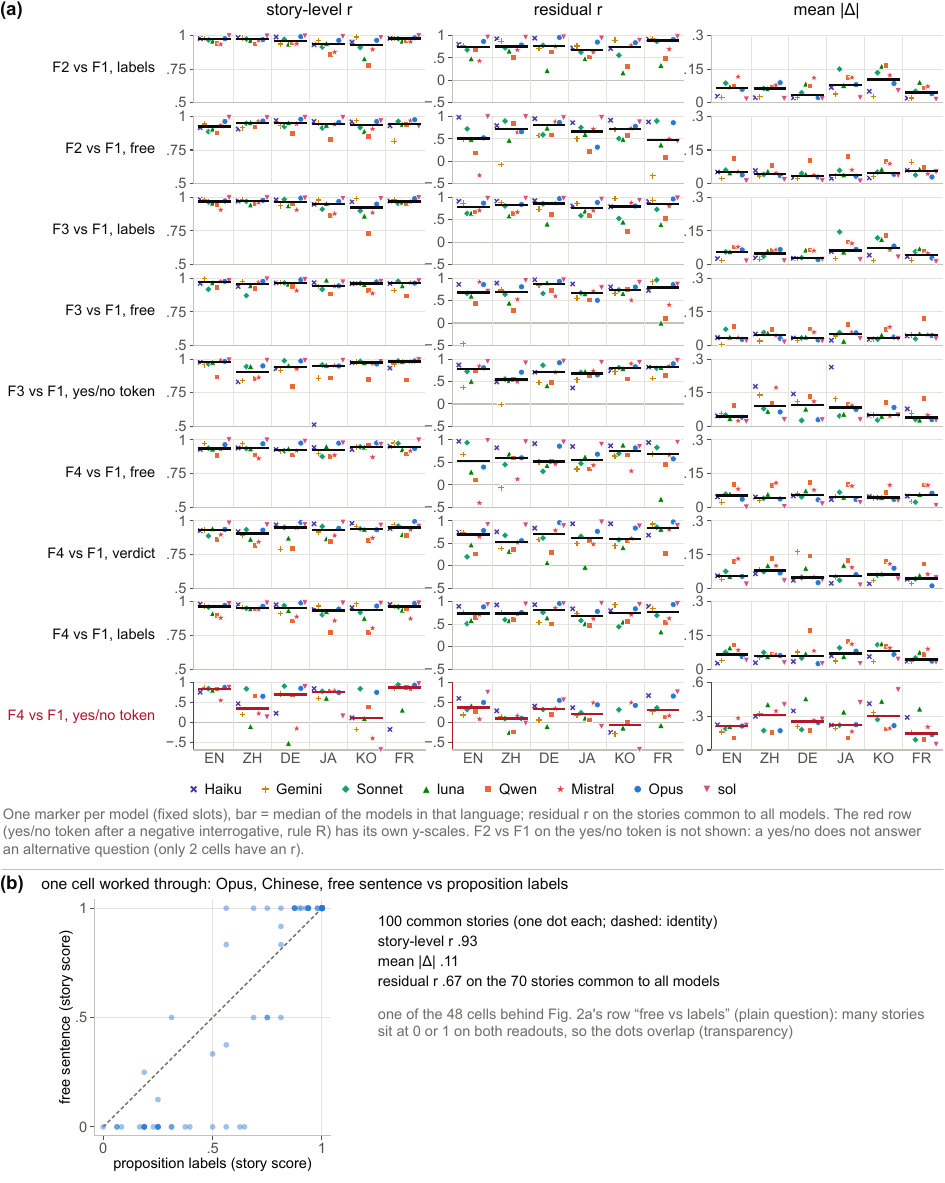}
\caption{Readouts and forms in detail. (a) Alternative, or-not, and negative-interrogative questions against plain questions, by answer constraint and language: per-cell story-level $r$, residual-profile $r$, and mean $|\Delta|$. (b) Opus in Chinese, free sentence versus proposition labels, on 100 common stories; the residual comparison uses the 70 stories common to all models.}
\label{fig:a2}
\end{figure*}

Figure~\ref{fig:a2} expands the form comparisons into individual model $\times$ language cells. Each marker compares story scores on the stories common to the two readouts; the bar summarises models within a language. Residual-profile correlations use the stories shared by all models in that language so that the story consensus has common support. The red row retains its own axis limits because the preset reading of the negative-interrogative token produces a wider spread. Table~\ref{tab:readouts} provides the cell counts, medians, ranges, and signed shifts for the comparisons summarised in the main text.

The worked example follows Opus in Chinese from story scores to their agreement statistics. Free sentences and proposition labels have a story-level correlation of $0.93$ on their common stories. Their mean absolute story difference is $0.11$, so close ordering coexists with changes in individual scores. Removing the story consensus and model means leaves a residual-profile correlation of $0.67$. The scatterplot displays many coincident endpoint scores with transparency, which prevents repeated points from appearing to be separate observations.

\setcounter{table}{2}
\begin{table*}[t]
\centering\footnotesize
\setlength{\tabcolsep}{3pt}
\begin{tabular}{@{}lrrrrr@{}}
\toprule
Readout comparison & Cells & Story-level $r$ & Residual $r$ & Mean $|\Delta|$ & Shift \\
\midrule
\multicolumn{6}{@{}l}{\textit{Plain question (F1): readout pairs}} \\
Free sentence vs proposition labels & $48$ & $0.92 [0.83, 0.98]$ & $0.49 [0.03, 0.87]$ & $0.10 [0.03, 0.17]$ & $-0.019$ \\
Free sentence vs verdict sentence & $48$ & $0.93 [0.88, 0.98]$ & $0.70 [0.05, 0.88]$ & $0.06 [0.03, 0.11]$ & $-0.024$ \\
Proposition labels vs verdict sentence & $48$ & $0.92 [0.84, 0.97]$ & $0.57 [0.26, 0.86]$ & $0.08 [0.04, 0.17]$ & $-0.002$ \\
Yes/no token vs proposition labels & $48$ & $0.93 [0.83, 0.97]$ & $0.49 [0.19, 0.87]$ & $0.10 [0.05, 0.17]$ & $-0.054$ \\
Yes/no token vs free sentence & $48$ & $0.92 [0.86, 0.98]$ & $0.50 [0.03, 0.85]$ & $0.08 [0.03, 0.18]$ & $-0.018$ \\
\addlinespace[2pt]
\multicolumn{6}{@{}l}{\textit{Alternative (F2) or or-not (F3) vs plain question (F1)}} \\
Alternative, proposition labels & $48$ & $0.96 [0.91, 0.99]$ & $0.76 [0.47, 0.94]$ & $0.07 [0.02, 0.12]$ & $-0.009$ \\
Or-not, proposition labels & $48$ & $0.97 [0.90, 0.99]$ & $0.81 [0.53, 0.97]$ & $0.06 [0.02, 0.10]$ & $-0.009$ \\
Alternative, free sentence & $48$ & $0.94 [0.88, 0.98]$ & $0.68 [0.21, 0.95]$ & $0.04 [0.02, 0.09]$ & $+0.019$ \\
Or-not, free sentence & $48$ & $0.96 [0.91, 0.98]$ & $0.73 [0.43, 0.92]$ & $0.04 [0.02, 0.07]$ & $-0.001$ \\
Or-not, yes/no token & $48$ & $0.96 [0.85, 0.99]$ & $0.73 [0.49, 0.92]$ & $0.06 [0.03, 0.14]$ & $-0.017$ \\
\addlinespace[2pt]
\multicolumn{6}{@{}l}{\textit{Negative interrogative (F4) vs plain question (F1)}} \\
Free sentence & $48$ & $0.94 [0.89, 0.98]$ & $0.66 [0.24, 0.94]$ & $0.05 [0.02, 0.10]$ & $-0.005$ \\
Verdict sentence & $48$ & $0.93 [0.85, 0.98]$ & $0.67 [0.29, 0.93]$ & $0.06 [0.02, 0.12]$ & $-0.021$ \\
Proposition labels & $48$ & $0.95 [0.87, 0.99]$ & $0.75 [0.51, 0.94]$ & $0.06 [0.03, 0.11]$ & $-0.007$ \\
Yes/no token & $48$ & $0.73 [-0.16, 0.89]$ & $0.22 [-0.17, 0.57]$ & $0.22 [0.13, 0.41]$ & $-0.095$ \\
\addlinespace[2pt]
\multicolumn{6}{@{}l}{\textit{Negative interrogative (F4): readout pairs}} \\
Yes/no token vs proposition labels & $48$ & $0.70 [-0.15, 0.89]$ & $0.24 [-0.10, 0.67]$ & $0.24 [0.14, 0.40]$ & $-0.136$ \\
Yes/no token vs free sentence & $48$ & $0.70 [-0.12, 0.89]$ & $0.19 [-0.18, 0.55]$ & $0.29 [0.19, 0.44]$ & $-0.131$ \\
Free sentence vs proposition labels & $48$ & $0.91 [0.84, 0.97]$ & $0.46 [0.10, 0.83]$ & $0.09 [0.05, 0.16]$ & $-0.028$ \\
\addlinespace[2pt]
\multicolumn{6}{@{}l}{\textit{Negative interrogative: judged yes/no-first replies}} \\
Judged sentence vs own token (same replies) & $48$ & $0.67 [-0.10, 0.94]$ & $0.30 [-0.18, 0.64]$ & $0.25 [0.12, 0.40]$ & $+0.106$ \\
Judged sentence vs free sentence (other replies) & $48$ & $0.94 [0.86, 0.97]$ & $0.61 [0.33, 0.87]$ & $0.07 [0.04, 0.13]$ & $-0.005$ \\
\addlinespace[2pt]
\bottomrule
\end{tabular}
\caption{Readout agreement on common stories within each cell. Entries are medians with the central 80\% range across cells; shift is the median of cell-wise mean signed differences, in the order named. Verdict sentence denotes the judged sentence of verdict-first replies. Same-reply estimates use identical replies and aggregation weights. The final row compares different answer conditions on common stories. The negative-interrogative token uses its preset mapping.}
\label{tab:readouts}
\end{table*}
 
\section{Further answer habits}
\label{app:habits}

\begin{figure*}[!t]
\centering
\includegraphics[width=\textwidth]{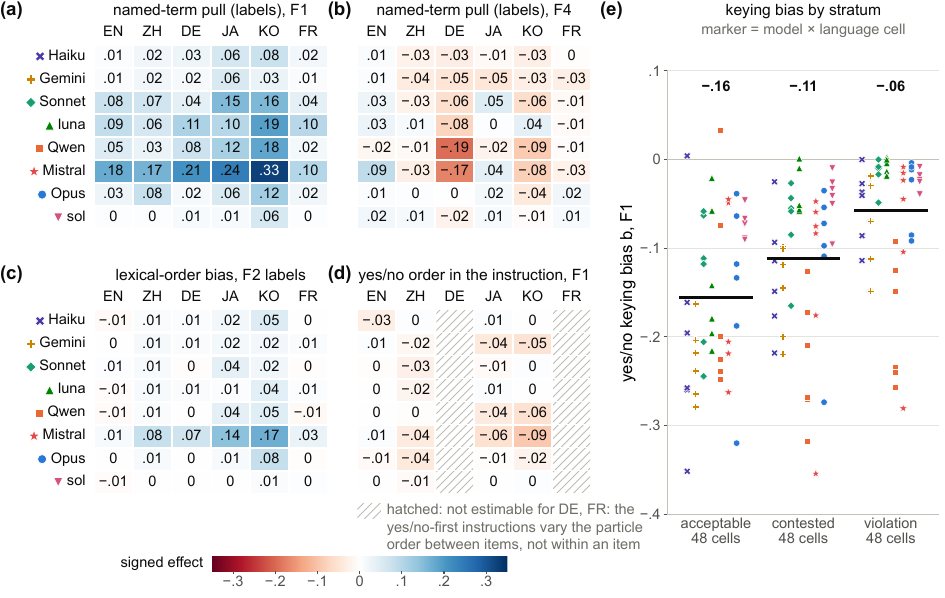}
\caption{Further answer habits. Named-term pull by model $\times$ language under plain questions and negative interrogatives, alternative-question lexical-order bias, particle-order effects in the yes/no-first instruction, and keying bias by story group. Particle-order effects are not estimable for German and French, whose instructions vary the first-listed particle between items. Small averages have local exceptions, especially Mistral.}
\label{fig:a3}
\end{figure*}

Figure~\ref{fig:a3} preserves the cell-level structure behind the language averages in Table~\ref{tab:habits}. Its named-term panels use the sign convention from the main text under both plain questions and negative interrogatives. The story-group panel compares the yes/no keying bias across acceptable, contested, and violation stories within each model $\times$ language cell. The negative keying bias is larger on acceptable stories than on violations, with contested stories between them.

The order panels distinguish the lexical order inside an alternative question from the particle order in the answer instruction. The particle-order effect is estimated within an item, holding its story, concept pair, named term, and question form fixed. German and French rotate the first-listed particle between items, so their corresponding cells are hatched as unavailable. Mistral supplies the clearest local departures from the small average order effects, particularly in the particle order of the Korean yes/no-first instruction and the lexical order of alternative questions.

\setcounter{table}{3}
\begin{table*}[t]
\centering\small
\setlength{\tabcolsep}{8pt}
\begin{tabular}{@{}lrrrrrr@{}}
\toprule
 & \multicolumn{2}{c}{Yes/no keying bias} & \multicolumn{2}{c}{Second-listed preference} & \multicolumn{2}{c}{Named-term pull} \\
Model & Plain & Negative & Plain & Negative & Plain & Negative \\
\midrule
Haiku & $-0.12$ & $-0.17$ & $+0.04$ & $+0.02$ & $+0.04$ & $-0.02$ \\
Gemini & $-0.14$ & $-0.04$ & $+0.01$ & $-0.01$ & $+0.02$ & $-0.03$ \\
Sonnet & $-0.07$ & $-0.04$ & $+0.11$ & $+0.08$ & $+0.09$ & $-0.01$ \\
luna & $-0.05$ & $+0.13$ & $+0.02$ & $-0.01$ & $+0.11$ & $+0.00$ \\
Qwen & $-0.19$ & $-0.16$ & $+0.05$ & $+0.07$ & $+0.08$ & $-0.06$ \\
Mistral & $-0.12$ & $-0.07$ & $+0.12$ & $+0.15$ & $+0.20$ & $-0.03$ \\
Opus & $-0.10$ & $-0.01$ & $+0.04$ & $+0.04$ & $+0.05$ & $+0.00$ \\
sol & $-0.05$ & $+0.06$ & $+0.02$ & $+0.01$ & $+0.02$ & $+0.00$ \\
\midrule
All cells & $-0.11$ & $-0.04$ & $+0.05$ & $+0.04$ & $+0.08$ & $-0.02$ \\
\bottomrule
\end{tabular}
\caption{Answer habits by model under plain questions and negative interrogatives, averaged equally over the six languages. The yes/no columns describe the opening token; the other columns use proposition labels. Figure~\ref{fig:habits} defines the directions, and Figure~\ref{fig:a3} gives the individual cells.}
\label{tab:habits}
\end{table*}
 
\section{The negative interrogative in detail}
\label{app:negation}

\begin{figure*}[p]
\centering
\includegraphics[width=\textwidth,height=.84\textheight,keepaspectratio]{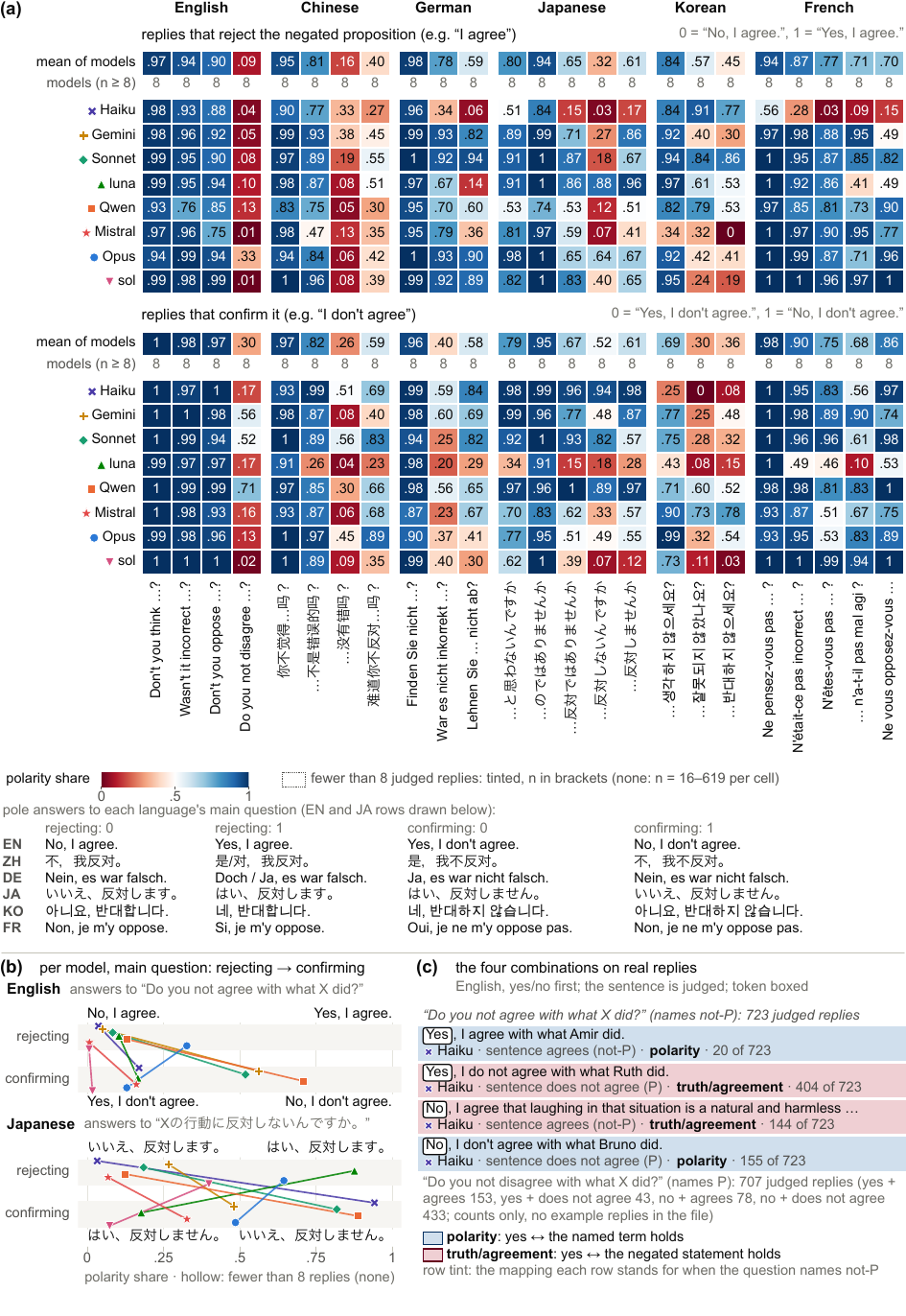}
\caption{The negative interrogative in detail. (a) Polarity share by wording and whether the reply confirms or rejects the negated proposition; each mean states its model count. (b) Per-model direction splits for ``Do you not agree\ldots?'' and \ja{反対しないんですか} (\emph{hantai shinai n desu ka}, ``Do you not oppose?''), with concrete answers at the poles. (c) The four particle--sentence combinations in replies to ``Do you not agree with what X did?'': one verbatim reply per combination, its model, and its count. Counts for ``Do you not disagree\ldots?'' appear separately.}
\label{fig:a4}
\end{figure*}

Figure~\ref{fig:a4} conditions the polarity share on what the judged sentence asserts. A rejecting sentence affirms the term questioned under negation, whereas a confirming sentence denies that term. The upper and lower matrices preserve the same wording rows and model columns, allowing a fixed opening particle to be distinguished from a mapping that follows the sentence. A model that always begins with ``no'' occupies opposite ends of the polarity scale across these directions.

Every displayed direction cell contains at least 15 usable replies, and every wording mean includes all eight models. The lower panels connect these aggregate patterns to actual sentences. The quoted examples all answer the question that names ``agree'': ``Do you not agree with what X did?''. Their counts establish that all four token--sentence combinations occur in the judged replies. Counts for the question naming ``disagree'' appear separately.

Table~\ref{tab:constructions} lists the prompt-defined construction groups used for the cross-language comparison. The table retains the construction-specific wording and each model's available reply count. A missing Japanese copula counterpart is left blank rather than replaced with a differently structured question.

\setcounter{table}{4}
\begin{table*}[t]
\centering\small
\setlength{\tabcolsep}{4pt}
\begin{tabularx}{\textwidth}{@{}l>{\raggedright\arraybackslash}Xrrrr@{}}
\toprule
Language & Wording & Mean & Model range & Replies & $n$ per model \\
\midrule
\multicolumn{6}{@{}l}{\textit{negated attitude verb}} \\
EN & Don't you think \ldots\ was wrong / right?; \ldots\ incorrect / correct? & 0.99 & 0.97--1.00 & 2\,799 & 282--382 \\
ZH & \zh{你不觉得…错了 / 对了吗；难道你不觉得…是错误的 / 正确的吗} & 0.96 & 0.92--1.00 & 2\,743 & 304--370 \\
DE & Finden Sie nicht, dass \ldots\ falsch / richtig war?; \ldots\ inkorrekt / korrekt? & 0.97 & 0.92--0.99 & 3\,136 & 355--408 \\
JA & \ja{…は間違っている / 正しいと思わないんですか} & 0.80 & 0.61--0.95 & 1\,611 & 192--210 \\
KO & \ko{…이 잘못됐다고 / 옳다고 생각하지 않으세요?; 틀렸다고 / 맞다고} & 0.77 & 0.57--0.96 & 2\,068 & 238--267 \\
FR & Ne pensez-vous pas que \ldots\ a mal / bien agi ? & 0.97 & 0.90--1.00 & 599 & 67--83 \\
\addlinespace
\multicolumn{6}{@{}l}{\textit{negated evaluative adjective}} \\
EN & Wasn't \ldots\ wrong / right?; Wasn't it incorrect / correct of \ldots? & 0.97 & 0.91--0.99 & 3\,208 & 354--429 \\
ZH & \zh{…没有错吗 / 不对吗；…不错误吗 / 不正确吗} & 0.21 & 0.06--0.41 & 1\,121 & 127--155 \\
DE & War \ldots\ nicht falsch / richtig?; War es nicht inkorrekt / korrekt \ldots? & 0.58 & 0.42--0.78 & 3\,203 & 378--410 \\
JA & No matched prompt in this inventory & --- & --- & --- & --- \\
KO & \ko{…은 잘못되지 / 옳지 않았나요?} & 0.41 & 0.18--0.67 & 632 & 70--86 \\
FR & N'était-ce pas incorrect / correct de la part de \ldots? & 0.91 & 0.71--1.00 & 1\,300 & 155--173 \\
\addlinespace
\multicolumn{6}{@{}l}{\textit{negated evaluative verb}} \\
EN & Do you not disagree / agree with \ldots?; oppose / support \ldots? & 0.22 & 0.01--0.50 & 2\,788 & 315--378 \\
ZH & \zh{难道你不反对 / 不同意…吗；难道你不反对 / 不支持…吗} & 0.47 & 0.29--0.70 & 6\,018 & 685--810 \\
DE & Lehnen Sie \ldots\ nicht ab? / Stimmen Sie \ldots\ nicht zu?; Lehnen Sie \ldots\ nicht ab? / Unterstützen Sie \ldots\ nicht? & 0.59 & 0.24--0.87 & 6\,359 & 752--824 \\
JA & \ja{…に反対 / 賛成しないんですか；…に反対 / を支持しないんですか} & 0.42 & 0.24--0.58 & 2\,618 & 290--344 \\
KO & \ko{…에 반대 / 찬성하지 않으세요?; …에 반대 / 을 지지하지 않으세요?} & 0.42 & 0.11--0.64 & 2\,775 & 337--357 \\
FR & N'êtes-vous pas en désaccord / d'accord avec \ldots?; Ne vous opposez-vous pas à \ldots? / Ne soutenez-vous pas \ldots? & 0.78 & 0.57--0.99 & 2\,661 & 311--357 \\
\addlinespace
\bottomrule
\end{tabularx}
\caption{Polarity share by corresponding construction, with an equal-weight mean over models and the range across models. The three families were defined from the prompts before their values were inspected; every member template is listed in the script of the prompt, glossed by its family name, with the two keyings of a pair separated by a slash and pairs by a semicolon. Counts are usable replies from the draw used in Figure~\ref{fig:negation}; every reported mean includes eight models. Where the language has a copula the adjective family negates copula and adjective (English contracted \emph{wasn't}); Chinese and Korean negate the evaluative predicate directly (Korean \ko{잘못되다}, ``be wrong''). The Chinese verb family carries the rhetorical adverb \zh{难道}. Language-specific forms without a counterpart stay out of the families and appear in Figure~\ref{fig:a4}: English contracted ``Don't you oppose \ldots?'', Chinese \zh{不是…吗} and \zh{难道…没有做错吗}, Japanese \ja{のではありませんか}, \ja{ませんか}, and \ja{ではありませんか}, French ``N'a-t-il pas mal agi?''.}
\label{tab:constructions}
\end{table*}
 
\section{Comparison with the published human reference}
\label{app:human}

\begin{figure*}[!t]
\centering
\includegraphics[width=\textwidth]{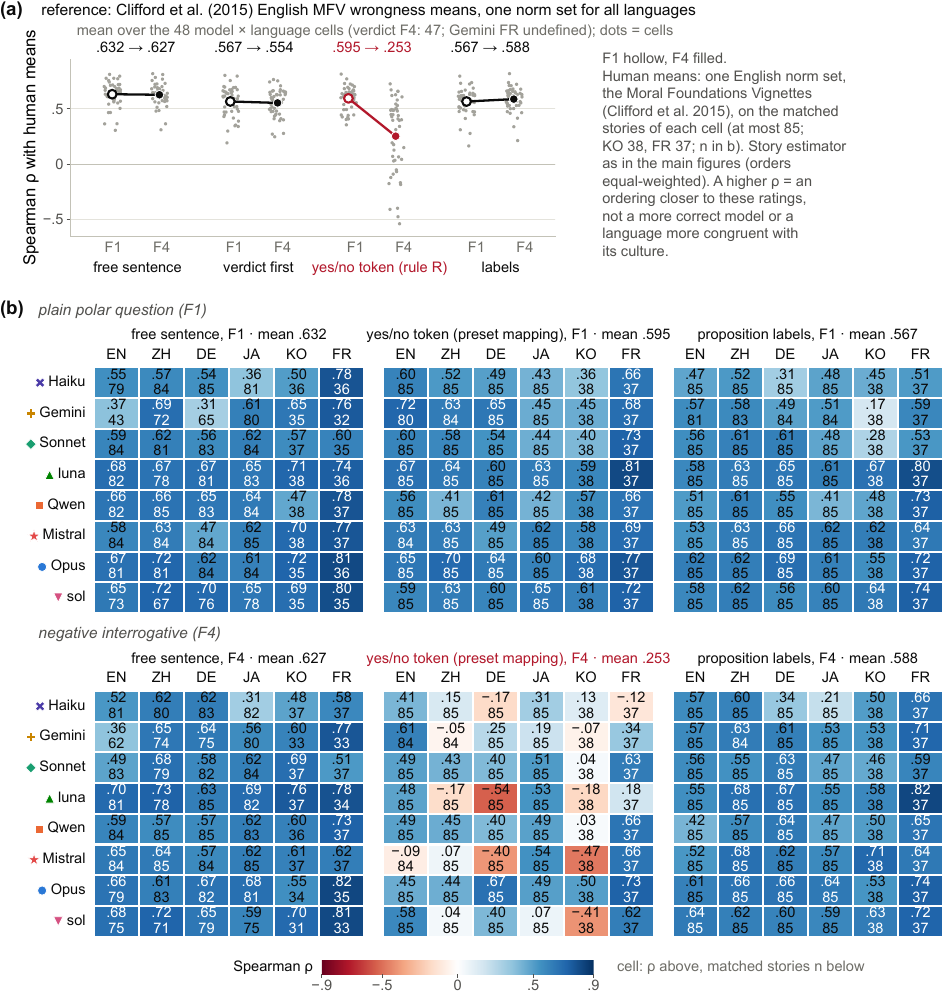}
\caption{Human reference. Spearman $\rho$ between the balanced story scores and published MFV wrongness means \citep{clifford2015}, using one English norm set for all languages and equal weighting of orders. (a) Mean $\rho$ by readout and form, with cell values. (b) Plain questions and negative interrogatives for free sentences, yes/no tokens under the preset mapping, and proposition labels, with $\rho$ above and matched-story count below. A higher correlation means an ordering closer to these ratings.}
\label{fig:a5}
\end{figure*}

We correlate the balanced story scores with the published English MFV wrongness means of \citet{clifford2015}. The same norm set is used for every language, restricted to stories present in the cell and readout being compared. Figure~\ref{fig:a5} prints each matched-story count below its Spearman correlation and uses the order-equal-weight estimator from the main analysis. The summary averages the available model $\times$ language correlations for each readout and question form.

The mean free-sentence correlation is $0.63$ under both the plain question and the negative interrogative. Proposition labels retain their ordering against the same reference under the form change. The preset token reading changes substantially, from $0.60$ to $0.25$. These correlations compare story rankings with a shared reference; their magnitude does not measure the cultural representativeness of a language or the correctness of a model's judgment.

\section{Dialogue-corpus and literature background}
\label{app:background}

\begin{figure*}[!t]
\centering
\includegraphics[width=\textwidth]{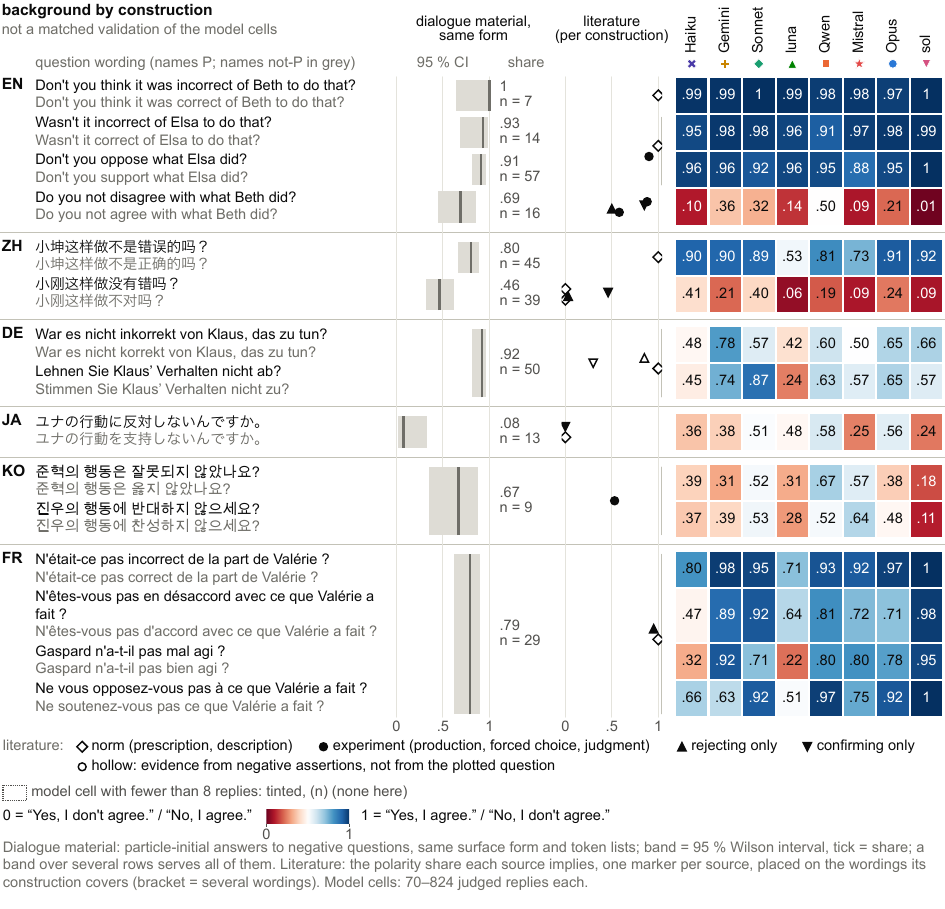}
\caption{Dialogue-corpus and literature background by construction. Particle-initial answers from transcribed conversation, subtitles, scenario-based meetings, crowdsourced dialogue, and institutional exchanges use the same surface forms and receive judge labels, with small samples. Literature anchors distinguish constructions; hollow experimental markers show evidence from negative assertions. The display supplies background for Figure~\ref{fig:negation} rather than a matched human validation of the model cells.}
\label{fig:a6}
\end{figure*}

Figure~\ref{fig:a6} places dialogue observations and published descriptions beside the battery's surface constructions as background for interpreting the particle mappings. The dialogue material contains particle-initial answers to negative questions whose contextual sentence interpretation was assigned by a judge. The particle lists follow the model-side lists, and the contextual label determines whether an answer confirms or rejects the negative proposition. The dialogue shares retain their sampling intervals and counts because construction coverage is uneven.

The sources include the Santa Barbara Corpus of Spoken American English \citep{dubois2005}, the AMI Meeting Corpus \citep{carletta2006}, CEFC--ORFEO \citep{benzitoun2016}, the Nagoya University Conversation Corpus \citep{fujimura2012}, and the BTSJ Japanese Natural Conversation Corpus \citep{usami2023}. The displayed construction groups also draw on Dortmund Chat \citep{lungen2016}, the Chinese dialogue collections LCCC and NaturalConv \citep{wang2020,wang2021}, OpenSubtitles \citep{lison2016}, and ParlaMint \citep{parlamint2025}. These resources include subtitles, scenario-based meetings, and crowdsourced dialogue, as well as transcribed conversation and chat.

Institutional exchanges come from UK Hansard through ParlParse, US Supreme Court oral arguments, and archived White House press briefings;\footnote{Source archives: \href{https://www.theyworkforyou.com/pwdata/scrapedxml/debates/}{Hansard/ParlParse}; \href{https://www.supremecourt.gov/oral_arguments/argument_transcripts/}{Supreme Court}; \href{https://bidenwhitehouse.archives.gov/briefing-room/press-briefings/}{White House}.} the French National Assembly, the Japanese National Diet, Korean government ministry briefings, and Taiwan's Legislative Yuan gazette through the g0v mirror.\footnote{Source endpoints: \href{https://data.assemblee-nationale.fr/static/openData/repository/17/vp/syceronbrut/syseron.xml.zip}{French National Assembly}; \href{https://kokkai.ndl.go.jp/api/speech}{National Diet}; \href{https://www.korea.kr/briefing/policyBriefingList.do}{Korean government}; \href{https://v2.ly.govapi.tw/gazette_agenda_doc/1154601_00001/html}{Legislative Yuan via g0v}.} The plotted construction groups determine which sources contribute; the figure does not represent a census of every response form in these corpora.

The literature markers distinguish descriptive accounts from experiments and distinguish rejecting answers from confirming answers when a source tests only one direction. The English anchors combine descriptions with experimental responses, elicited \citep{goodhue2018} or selected from a given yes/no pair \citep{park2023}, and theoretical accounts \citep{kramer2011,kameyama2013}; the Mandarin anchors combine descriptive accounts with elicited production \citep{lu2002,holmberg2016,zhang2021}. Japanese and French markers draw on descriptions and elicited responses \citep{kuno1973,kameyama2013}, and the Korean marker uses forced-choice answer selection \citep{park2023}. German anchors include a grammatical description \citep{duden2009} and acceptability judgments about negative assertions \citep{meijer2015,claus2017}. Hollow experimental markers identify these assertion studies, which concern a related discourse context rather than the plotted question construction.

Keeping the particle form separate from contextual confirmation also follows the distinct coding categories of \citet{stivers2010}; \mbox{\citet{enfield2019}} analyse form choice within confirming answers. Mandarin conversation studies likewise distinguish single-word and repetition-based answers and show that the negative expression \zh{没有} (\emph{méiyǒu}) can do different interactional work depending on the preceding question \citep{xie2018,wangwei2020,wangyufang2008}. The dialogue exchanges differ from the vignette prompts in topic, register, and conversational context, and provide background rather than a matched validation of the model cells.

\section{Concept pairs and wording sensitivity}
\label{app:pairs}

\begin{figure*}[!t]
\centering
\includegraphics[width=\textwidth]{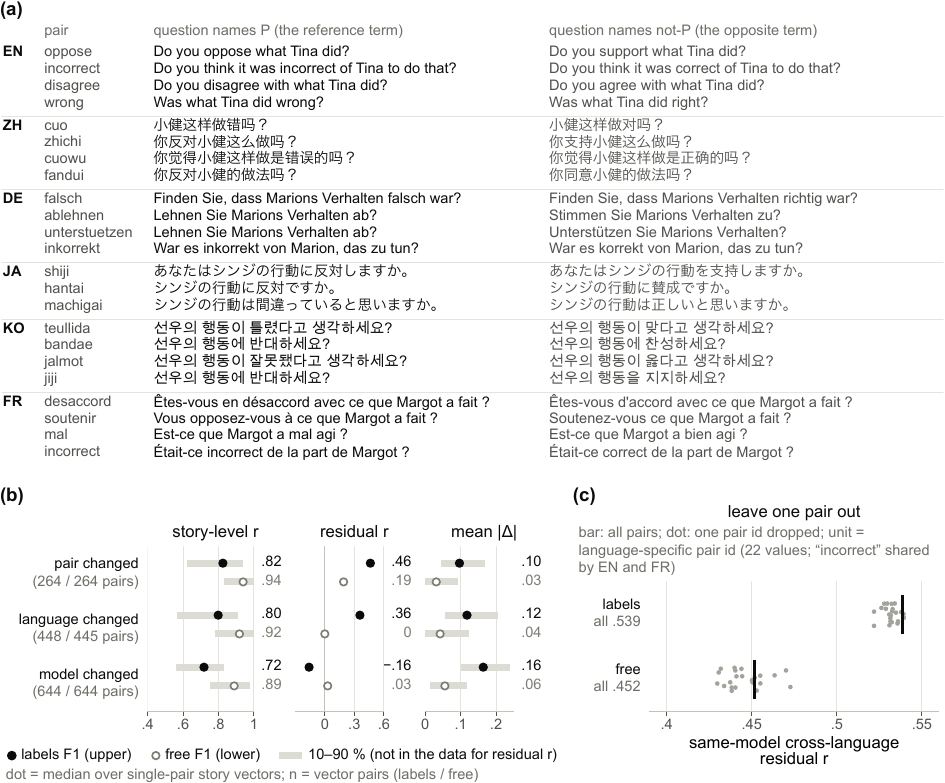}
\caption{Concept pairs. Questions naming either term of the 23 language-specific pairs; wording sensitivity when changing pair, language, or model on single-pair story vectors; and leave-one-pair-out stability of the same-model cross-language residual correlation. Orders are equally weighted. The leave-one-pair-out analysis has 22 units because the English and French pairs named \emph{incorrect} are removed together.}
\label{fig:a7}
\end{figure*}

Figure~\ref{fig:a7} lists the questions naming either term of every language-specific concept pair. The pair identifies the evaluative contrast, and each question names one of its terms. The sensitivity analysis changes one of pair, language, or model while comparing single-pair story vectors on their common stories. It therefore asks how much a wording choice changes the measurement before the story estimator averages over pairs.

For proposition labels, changing the concept pair within a model and language gives a median story-level correlation of $0.82$. Changing the model within a language and pair gives $0.72$. The corresponding free-sentence correlations are higher, but their residual-profile correlations reveal less shared variation after the story consensus is removed. Averaging concept pairs consequently has a distinct measurement purpose alongside balancing which term of each pair the question names.

The leave-one-pair-out analysis repeats the same-model cross-language residual comparison after excluding each concept-pair name in turn. The inventory contains 23 language-specific entries but 22 exclusions because the English and French pairs named \emph{incorrect} are removed together. The figure displays every exclusion beside the all-pairs reference so the effect of each exclusion remains visible.

\section{Repetition and aggregation stability}
\label{app:stability}

\begin{figure*}[!t]
\centering
\includegraphics[width=\textwidth]{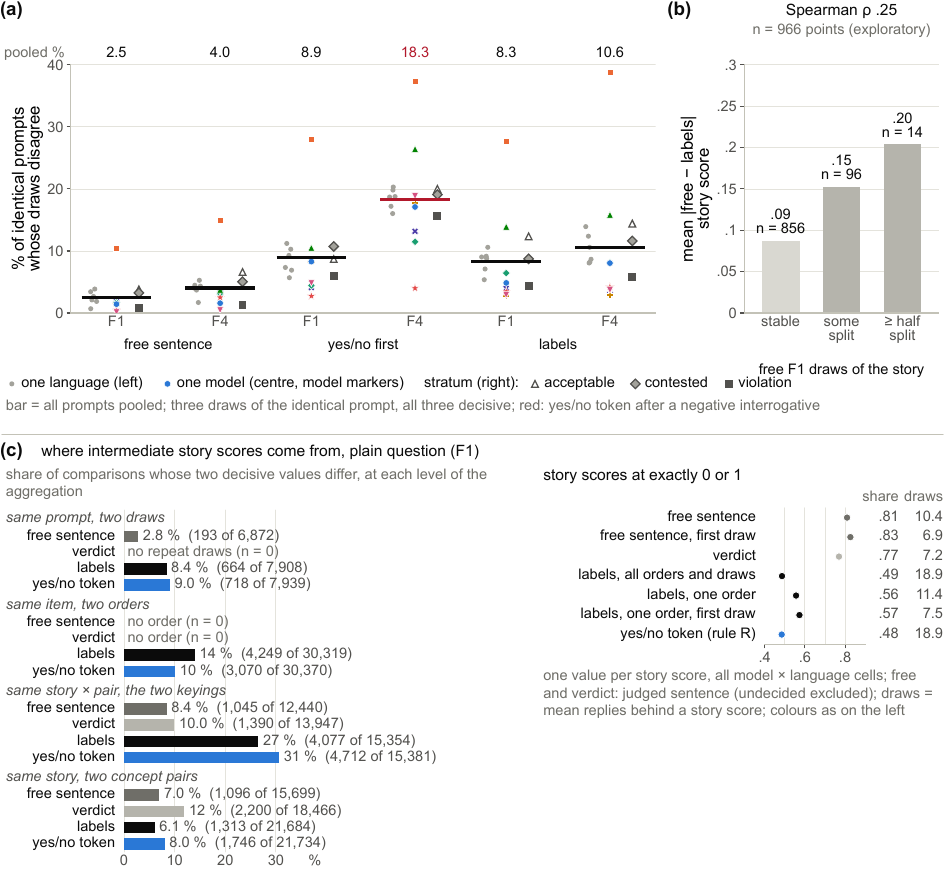}
\caption{Repetition and aggregation stability. (a) Share of identical prompts whose three decisive generations disagree, by answer constraint, question form, language, model, and story group. (b) Stories with unstable generations also disagree more between readouts; the relationship is exploratory and uses shared sampling. (c) Sources of intermediate story scores under plain questions: repeated generations, option orders, named terms, and concept pairs, followed by the share of story scores at either endpoint.}
\label{fig:a8}
\end{figure*}

Figure~\ref{fig:a8} separates variation across repeated generations from variation introduced when scores average across option orders, terms named in the question, and concept pairs. A repeated prompt fixes the model, language, story, concept pair, question form, named term, answer instruction, and option order. The repetition panel uses prompts with three decisive generations and records whether those readings split. For free sentences, the split share rises from $2.5\%$ under plain questions to $4.0\%$ under negative interrogatives. The pattern differs across models and answer constraints, as the markers in the figure make visible.

The association between unstable repetitions and disagreement across readouts is exploratory because both quantities use overlapping sampled replies. Across eligible model $\times$ language $\times$ story observations, the Spearman correlation is $0.25$. This association locates stories on which repeated sampling contributes to readout variability without assigning that variability to a particular cause.

The aggregation audit follows the successive sources of intermediate scores. A difference across repeated generations or label orders is an observed change in the decisive reading. For changes of the named term or concept pair, the audit records differences of at least $0.5$ after the lower-level averages. Label choices differ more often when the named term changes than free-sentence judgments, $27\%$ versus $8\%$. Order changes contribute further intermediate label scores. The panel reports the denominator at each aggregation level so these shares can be read as different comparisons.

Restricting label aggregation reduces, but leaves, the endpoint difference. Free-sentence scores are at an endpoint in $81\%$ of model $\times$ language $\times$ story observations, against $57\%$ for labels restricted to one order and one generation. The judged examples include hedged answers classified as taking both sides or conditions, while counting undecided sentences as the midpoint does not improve agreement with labels. The audit therefore retains the decisive-only estimator used throughout the paper.

\section{Mean-level views and consistency across readouts}
\label{app:means}

\begin{figure*}[!t]
\centering
\includegraphics[width=\textwidth]{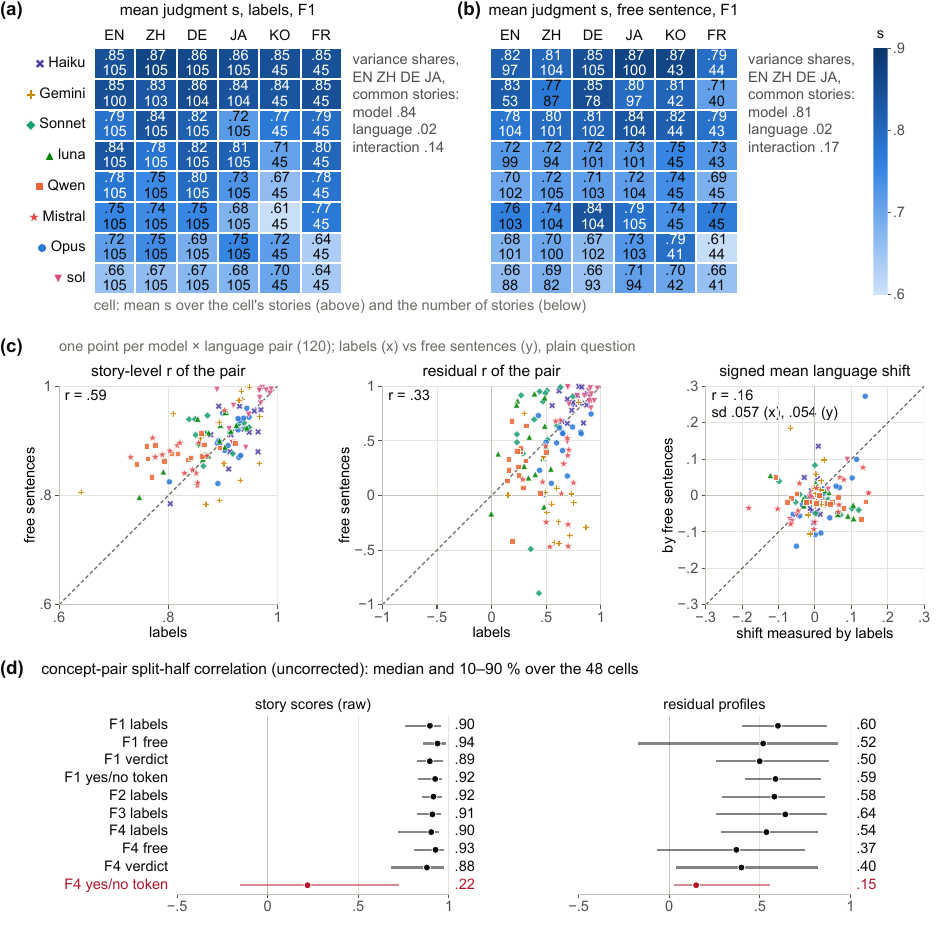}
\caption{Mean-level views. Cell-mean heat maps with story counts; common-story variance shares for the four languages with broader coverage; signed mean language shifts under labels and free sentences; and replication of each model's language-pair correlations across those readouts. The final panel gives concept-pair split-half correlations, uncorrected, as a reference for repeated measurement within a readout.}
\label{fig:a9}
\end{figure*}

Figure~\ref{fig:a9} complements the pairwise profile comparison with cell means and replication across readouts. The heat maps average each cell's available stories and print that support beneath the mean. The variance decomposition instead fixes a common story set across models and the four languages with the broader coverage. Table~\ref{tab:pairs} records the pairwise class estimates and the fixed-support sensitivity check, preserving their different units of aggregation.

For English, Chinese, German, and Japanese, model differences account for $0.84$ of the proposition-label mean matrix's variance. The corresponding share for free sentences is $0.81$. Language main effects are small in these common-story matrices, while the interaction retains differences specific to a model and language. These mean-level summaries complement the story-profile comparison because averaging over stories removes the item structure that identifies a model.

The replication panels use one observation for each model and language pair. The cross-language story-level correlations under labels and free sentences correlate at $0.59$; the residual-profile correlations correlate at $0.33$. The signed mean language shifts reproduce less closely across these readouts. The final panel splits each cell's concept pairs into two subsets (two and two, or one and two for Japanese, which has three pairs) and reports the correlation between the two subsets' story profiles, uncorrected, as a reference for how much profile structure repeats within each readout.

Table~\ref{tab:pairs} adds two readings of the same cell structure. The nearest-neighbor block treats each cell as an unlabeled profile: under residual $r$, 39 of the 48 proposition-label cells have another language of the same model as their closest other cell, against 5 expected by chance, and the five same-model cells rank on average 5.7th among a cell's 47 neighbors. Under free sentences the count falls to 15 of 48. The story-group block recomputes the class means on acceptable, contested, and violation stories separately, over pairs with at least eight common stories of that group. The same-model advantage in story-level correlation is largest on acceptable stories ($0.79$ against $0.69$), present on contested stories ($0.70$ against $0.62$), and absent on violations ($0.43$ against $0.46$), where every cell condemns the act; the mean absolute story difference and the residual correlation keep the same-model advantage in every group. The overall story-level correlations of $0.88$ and $0.81$ are therefore raised by the structure that every cell shares across groups, while the model signature lives in the stories on which models disagree.

\setcounter{table}{5}
\begin{table*}[!t]
\centering\footnotesize
\setlength{\tabcolsep}{3pt}
\begin{tabular}{@{}lrrrr@{}}
\toprule
Class of cell pairs & Pairs & Story-level $r$ & Residual $r$ & Mean $|\Delta|$ \\
\midrule
\multicolumn{5}{@{}l}{\textit{Proposition labels}} \\
Same model, different language & 120 & $0.88 [0.83, 0.91]$ & $0.56 [0.51, 0.60]$ & $0.10 [0.09, 0.11]$ \\
Same language, different model & 168 & $0.81 [0.75, 0.86]$ & $-0.12 [-0.13, -0.11]$ & $0.15 [0.13, 0.17]$ \\
Both different & 840 & $0.79 [0.71, 0.84]$ & $-0.06 [-0.07, -0.05]$ & $0.16 [0.14, 0.18]$ \\
Class difference & --- & $+0.075 [+0.047, +0.096]$ & $+0.677 [+0.627, +0.718]$ & $-0.052 [-0.065, -0.040]$ \\
\addlinespace
\multicolumn{5}{@{}l}{\textit{Free sentence}} \\
Same model, different language & 120 & $0.90 [0.86, 0.93]$ & $0.45 [0.22, 0.52]$ & $0.07 [0.05, 0.09]$ \\
Same language, different model & 168 & $0.87 [0.82, 0.90]$ & $-0.12 [-0.12, -0.02]$ & $0.09 [0.07, 0.12]$ \\
Both different & 840 & $0.84 [0.78, 0.89]$ & $-0.06 [-0.06, 0.02]$ & $0.11 [0.07, 0.14]$ \\
Class difference & --- & $+0.037 [+0.011, +0.068]$ & $+0.573 [+0.281, +0.629]$ & $-0.020 [-0.035, -0.008]$ \\
\addlinespace
\bottomrule
\end{tabular}
\par\medskip
\begin{tabular}{@{}llrrrrrr@{}}
\toprule
 & & & & \multicolumn{2}{c}{Story-level $r$} & \multicolumn{2}{c}{Mean $|\Delta|$} \\
Readout & Languages & Stories & Cells & Same model & Same language & Same model & Same language \\
\midrule
Proposition labels & EN/ZH/DE/JA & $100$ & $32$ & $0.907$ & $0.832$ & $0.078$ & $0.133$ \\
 & All languages & $17$ & $48$ & $0.844$ & $0.734$ & $0.116$ & $0.190$ \\
Free sentence & EN/ZH/DE/JA & $46$ & $32$ & $0.981$ & $0.972$ & $0.015$ & $0.020$ \\
\bottomrule
\end{tabular}
\par\medskip
\begin{tabular}{@{}lrrrrrr@{}}
\toprule
 & \multicolumn{3}{c}{Nearest cell is the same model (of 48; chance 5)} & \multicolumn{3}{c}{Mean rank of same-model cells (of 47; chance 24)} \\
Readout & Story $r$ & Residual $r$ & Mean $|\Delta|$ & Story $r$ & Residual $r$ & Mean $|\Delta|$ \\
\midrule
Proposition labels & $25$ & $39$ & $30$ & $9.9$ & $5.7$ & $9.9$ \\
Free sentence & $12$ & $15$ & $10$ & $13.7$ & $12.8$ & $13.5$ \\
\bottomrule
\end{tabular}
\par\medskip
\begin{tabular}{@{}lrrrrrr@{}}
\toprule
 & \multicolumn{2}{c}{Story-level $r$} & \multicolumn{2}{c}{Mean $|\Delta|$} & \multicolumn{2}{c}{Residual $r$} \\
Story group (proposition labels) & Same model & Same language & Same model & Same language & Same model & Same language \\
\midrule
Acceptable & $0.79$ & $0.69$ & $0.14$ & $0.25$ & $0.54$ & $-0.13$ \\
Contested & $0.70$ & $0.62$ & $0.09$ & $0.14$ & $0.43$ & $-0.09$ \\
Violation & $0.43$ & $0.46$ & $0.05$ & $0.06$ & $0.36$ & $-0.10$ \\
\bottomrule
\end{tabular}
\caption{Cell-pair classes for plain questions. The first block gives class means and 95\% story-bootstrap intervals on pairwise-common stories; differences subtract the same-language class from the same-model class and use a joint bootstrap. The second block fixes the story set across every included cell; the free-sentence readout has only seven stories common to all languages, so that estimate is omitted. The third block treats each cell as an unlabeled profile and asks which other cell it resembles most, and where its five same-model cells rank among its 47 neighbors. The fourth block recomputes the class means on the stories of one group, over pairs with at least eight common stories of that group. Models and languages are fixed. Within-language residual centring constrains the same-language residual class.}
\label{tab:pairs}
\end{table*}
 
\section{Parsing audit}
\label{app:parsing}

\begin{figure*}[p]
\centering
\includegraphics[width=\textwidth,height=.80\textheight,keepaspectratio]{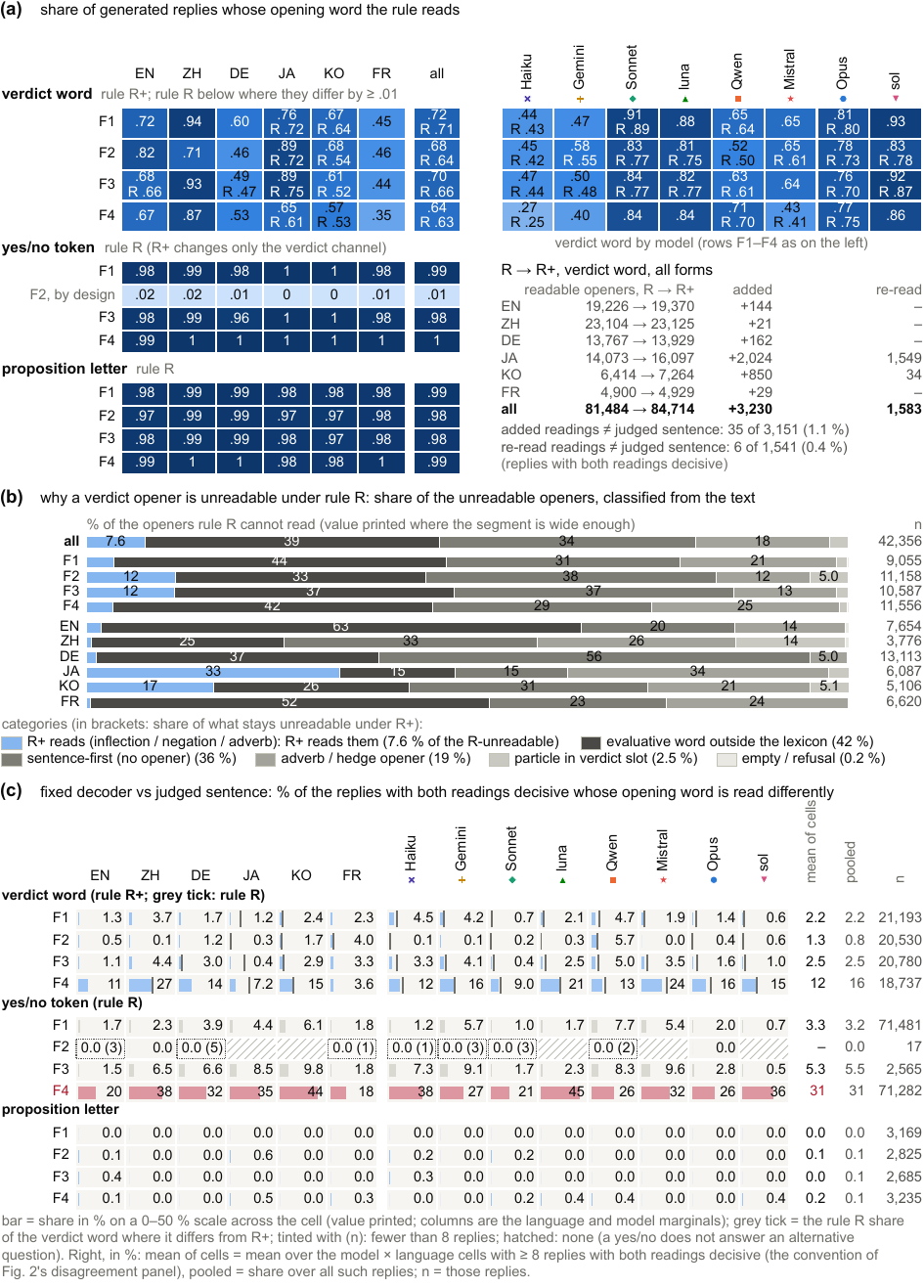}
\caption{Parsing audit: the fixed decoder against the sentence judge. (a) Opener readability by answer constraint, question form, language, and model under the fixed verdict-word rule and its extension. (b) Causes of unreadable verdict openings, with the portion recovered by the extension shown separately. (c) Readout disagreement on replies where both readings are decisive, by question form and language or model: the verdict word under the extended rule, with grey ticks for the original rule where it differs, the token and the letter under the fixed rule. The summary columns give the mean over model $\times$ language cells, the pooled share, and $n$. Of the 71,282 negative-interrogative token replies, 21 opened with a verdict word rather than a particle (20 Chinese, 1 Japanese). Replies behind the series of Figure~\ref{fig:convergence}b: yes/no token 71,481 / 2,565 / 71,282 (F1 / F3 / F4; the token after an alternative question has 17 such replies and is not plotted); verdict word 21,193 / 20,530 / 20,780 / 18,737 (F1--F4); letter 3,169 / 2,825 / 2,685 / 3,235.}
\label{fig:a10}
\end{figure*}

Figure~\ref{fig:a10} distinguishes an unreadable opening from a disagreement between a readable opening and the judged sentence. Readability is measured over generated replies. Disagreement is measured over replies for which both readings are decisive, with equal weight assigned to model $\times$ language cells in the figure's summary. Table~\ref{tab:audits} retains the reply denominators alongside the judging audits.

The extension to the fixed verdict-word rule consists of four stated operations. It recognises inflected and polite forms of listed words, reads preposed negation immediately before a listed term, recognises derived adverbs, and reverses a previously readable term when the following negation changes its polarity. For example, \ja{反対しません} (\emph{hantai shimasen}, ``I do not oppose'') is read as a negated verdict instead of the unnegated \ja{反対} (\emph{hantai}, ``oppose''). Evaluative words outside the listed vocabulary and replies that begin with a full sentence remain for the sentence judge.

The extension raises overall verdict-opener readability from $0.66$ to $0.68$. It also changes readings that the original prefix rule had assigned to unnegated terms. Among Japanese plain-question verdict replies, the resulting readout disagreement decreases from $12.5\%$ to $1.2\%$. The before-and-after values in Figure~\ref{fig:a10} distinguish these changes in reading from the additional replies that become readable.

The reason categories describe the openings left unreadable by the rule. Figure~\ref{fig:a10}b starts from the original unreadable pool and identifies the portion recovered by the extension. Its bracketed legend reports the composition of the remaining unreadable pool under the extended rule. The bar segments use the original unreadable pool as their denominator, whereas the legend describes what remains unreadable after the extension.

\setcounter{table}{6}
\begin{table*}[!t]
\centering\small
\setlength{\tabcolsep}{4pt}
\begin{tabular}{@{}lrrrrr@{}}
\toprule
Audit against the primary sentence judge & Replies & Agreement & Interval & $\kappa$ & Decisive-ref. $n$ \\
\midrule
GPT-6 astra, high & 4\,590 & 95.5\% & 94.9--96.1\% & 0.90 & 4\,224 \\
Claude Fable 5, high & 4\,530 & 97.2\% & 96.7--97.7\% & 0.94 & 4\,168 \\
\shortstack[l]{Same judge, repeated run\\on identical prompts (set 1)} & 300 & 96.3\% & --- & 0.92 & --- \\
\shortstack[l]{Same judge, repeated run\\on identical prompts (set 2)} & 300 & 97.7\% & --- & 0.95 & --- \\
\shortstack[l]{Native readers, 100 replies per language\\(Chinese, English, German, Japanese)} & 400 & 95.5\% & 93.0--97.1\% & 0.89 & 363 \\
\bottomrule
\end{tabular}
\par\medskip
\begin{tabular}{@{}lrrrrrr@{}}
\toprule
 & \multicolumn{3}{c}{GPT-6 astra, high} & \multicolumn{3}{c}{Claude Fable 5, high} \\
Stratum of the audited replies & $n$ & Agreement & Opposite & $n$ & Agreement & Opposite \\
\midrule
All audited replies & 4\,590 & 95.5\% & 10 & 4\,530 & 97.2\% & 9 \\
Negative interrogative, all constraints & 1\,225 & 95.1\% & 7 & 1\,209 & 96.7\% & 5 \\
\quad free sentence & 394 & 94.7\% & 1 & 394 & 95.7\% & 1 \\
\quad verdict first & 407 & 95.3\% & 2 & 407 & 97.1\% & 1 \\
\quad proposition labels & 151 & 98.0\% & 2 & 149 & 98.7\% & 2 \\
\quad yes/no first & 152 & 93.4\% & 2 & 143 & 96.5\% & 1 \\
\quad\quad fixed rule and sentence disagree & 47 & 93.6\% & 2 & 43 & 97.7\% & 1 \\
\bottomrule
\end{tabular}
\par\medskip
\begin{tabular}{@{}lrrrrrr@{}}
\toprule
 & \multicolumn{2}{c}{Claude Fable 5, high} & \multicolumn{2}{c}{Claude Opus 5.5, high} & \multicolumn{2}{c}{Claude Sonnet 5, high} \\
Core strata (rule and primary label both decisive) & $n$ & Agreement & $n$ & Agreement & $n$ & Agreement \\
\midrule
Negative interrogative, yes/no first: readings agree & 955 & 96.5\% & 960 & 96.1\% & 903 & 98.0\% \\
\quad readings conflict & 938 & 93.5\% & 953 & 92.8\% & 880 & 96.4\% \\
Plain question, yes/no first: readings agree & 475 & 98.9\% & 480 & 99.0\% & 457 & 98.9\% \\
\quad readings conflict & 437 & 48.7\% & 447 & 49.0\% & 432 & 48.1\% \\
Negative interrogative, verdict first (R+): agree & 235 & 97.0\% & 240 & 97.5\% & 220 & 97.7\% \\
\quad readings conflict & 178 & 91.0\% & 183 & 90.7\% & 165 & 90.9\% \\
\midrule
All core replies & 3\,218 & 89.2\% & 3\,263 & 88.9\% & 3\,057 & 90.2\% \\
\bottomrule
\end{tabular}
\par\medskip
\begin{tabular}{@{}llrrrrrr@{}}
\toprule
 & & & & & \multicolumn{2}{c}{Readable opener (\%)} & Both decisive \\
Constraint & Form & Generated & Judged & Sentence decisive & Fixed rule & Extended rule & Extended rule \\
\midrule
Verdict first & F1 & 30\,960 & 30\,960 & 29\,023 & 70.8 & 71.8 & 21\,193 \\
 & F2 & 30\,960 & 30\,960 & 29\,430 & 64.0 & 68.2 & 20\,530 \\
 & F3 & 30\,960 & 30\,960 & 29\,323 & 65.8 & 69.7 & 20\,780 \\
 & F4 & 30\,960 & 30\,960 & 28\,590 & 62.7 & 64.0 & 18\,737 \\
\addlinespace
Yes/no first & F1 & 77\,952 & 77\,952 & 71\,883 & 98.8 & 98.8 & 71\,481 \\
 & F2 & 61\,920 & 2\,737 & 2\,554 & 1.0 & 1.0 & 17 \\
 & F3 & 61\,920 & 2\,728 & 2\,585 & 98.2 & 98.2 & 2\,565 \\
 & F4 & 77\,952 & 77\,952 & 71\,291 & 99.8 & 99.8 & 71\,282 \\
\addlinespace
Proposition labels & F1 & 77\,951 & 3\,211 & 3\,184 & 98.6 & 98.6 & 3\,169 \\
 & F2 & 61\,920 & 2\,882 & 2\,838 & 98.0 & 98.0 & 2\,825 \\
 & F3 & 61\,919 & 2\,733 & 2\,691 & 98.2 & 98.2 & 2\,685 \\
 & F4 & 77\,952 & 3\,253 & 3\,247 & 99.2 & 99.2 & 3\,235 \\
\addlinespace
\bottomrule
\end{tabular}
\caption{Judge audits and opener coverage. Audit intervals are 95\% intervals; each test--retest set repeats the judge on 300 negative-interrogative yes/no-first replies, 50 per language, with one and three flips between the reference proposition and its opposite. The native readers' labels agree with the astra labels on 93.0\% and with the Fable labels on 95.4\% of the same replies, and no decisive reader label opposes a decisive machine label. The stratum block gives agreement with the primary judge and the number of replies on which the two judges assert opposite decisive propositions (``Opposite''); the audited negative-interrogative yes/no-first replies number 19--33 per language. The core-strata block re-judges, with three Claude judge configurations at high reasoning effort, a seeded sample of replies on which the fixed rule and the primary sentence label are both decisive, drawn per model $\times$ language cell and split by whether the two readings agree or conflict; agreement is with the primary label, so on conflict rows it is the share of replies on which the audit judge upholds the sentence reading against the fixed rule. Over all audited replies, the remaining disagreements move a reply between a decisive and an undecided category (134 and 42 for astra; 47 and 54 for Fable) or between the two undecided categories (19; 15). Coverage includes the three constraints shown, excluding a 13\,824-reply control stratum. Readability divides readable openers by generated replies. The extension changes the verdict-word rule; alternative-question particles are unreadable by design. Both decisive requires a decisive sentence and a readable opener. Forms: F1 plain polar, F2 alternative, F3 or-not, F4 negative interrogative.}
\label{tab:audits}
\end{table*}
 
 \end{document}